\documentclass[accepted]{uai2023} 
\pdfoutput=1

\usepackage[american]{babel}

\usepackage[round]{natbib} 
\usepackage{amsmath,amsfonts,bm}

\def\eqref#1{equation~\ref{#1}}

\def\1{\bm{1}}

\def\vzero{{\bm{0}}}

\def\va{{\bm{a}}}
\def\vb{{\bm{b}}}

\def\vd{{\bm{d}}}
\def\ve{{\bm{e}}}
\def\vf{{\bm{f}}}
\def\vg{{\bm{g}}}
\def\vh{{\bm{h}}}

\def\vm{{\bm{m}}}

\def\vvv{{\bm{v}}}
\def\vw{{\bm{w}}}
\def\vx{{\bm{x}}}

\def\vz{{\bm{z}}}

\def\mD{{\bm{D}}}

\def\mH{{\bm{H}}}
\def\mI{{\bm{I}}}
\def\mJ{{\bm{J}}}

\def\mM{{\bm{M}}}

\def\mP{{\bm{P}}}
\def\mQ{{\bm{Q}}}

\def\mS{{\bm{S}}}

\def\mU{{\bm{U}}}
\def\mV{{\bm{V}}}

\DeclareMathAlphabet{\mathsfit}{\encodingdefault}{\sfdefault}{m}{sl}
\SetMathAlphabet{\mathsfit}{bold}{\encodingdefault}{\sfdefault}{bx}{n}

\usepackage[dvipsnames]{xcolor}         
\usepackage{hyperref}
\hypersetup{
colorlinks = true,
linkcolor = ForestGreen,
anchorcolor = blue,
citecolor = Blue,
filecolor = cyan,
menucolor = ForestGreen,
runcolor = cyan,
urlcolor = ForestGreen}

\usepackage{url}
\usepackage{booktabs}       
\usepackage{amsfonts}       
\usepackage{amssymb}
\usepackage{nicefrac}       
\usepackage{microtype}      
\usepackage{multicol, multirow}

\usepackage{amsmath, mathtools, amssymb}
\usepackage[capitalise]{cleveref}

\usepackage{subcaption, caption}
\usepackage{bm}
\usepackage{adjustbox}
\usepackage{tikz}
\usepackage{wrapfig}

\usetikzlibrary{calc,matrix,positioning,patterns,shapes.geometric}

\DeclareMathOperator{\diag}{diag}

\DeclareMathOperator{\Covv}{Cov}
\newtheorem{definition}{Definition}[section]
\newtheorem{lemma}{Lemma}[section]

\newtheorem{theorem}{Theorem}[section]

\crefname{condition}{condition}{conditions}
\Crefname{condition}{Condition}{Conditions}
\crefname{example}{example}{example}
\Crefname{example}{Example}{Example}
\Crefname{section}{Section}{Section} 
\crefname{section}{Sec.}{Sec.} 
\crefname{figure}{Fig.}{Figs.} 
\Crefname{figure}{Figure}{Figures} 
\Crefname{table}{Table}{Tables} 
\crefname{table}{Tab.}{Tab.} 
\Crefname{equation}{Equation}{Equations} 
\crefname{equation}{Eqn.}{Eqns.} 
\Crefname{algocf}{Algorithm}{Algorithms} 
\crefname{algocf}{Alg.}{Algs.} 
\Crefname{theorem}{Theorem}{Theorem} 
\crefname{theorem}{Thm.}{Thms.} 

\newcommand{\jacf}[0]{\bm{J}_{f}}

\newcommand{\jacfstar}[0]{\bm{J}_{f^*}}
\newcommand{\jacfprime}[0]{\bm{J}_{f^{'}}}

\usepackage{tikz}
\usepackage{esvect}

\usepackage[ruled]{algorithm2e}

\newcommand{\added}[1]{#1}

\title{When are Post-hoc Conceptual Explanations Identifiable?}

\author[1,2,$\dagger$]{\href{mailto:tobias.leemann@uni-tuebingen.de}{\textcolor{black}{Tobias Leemann}}{}}
\author[1,$\dagger$]{\href{mailto:michael.kirchhof@uni-tuebingen.de}{\textcolor{black}{Michael Kirchhof}}{}}
\author[1,2]{Yao Rong}
\author[2]{Enkelejda Kasneci}
\author[2]{Gjergji Kasneci}
\affil[1]{%
    University of Tübingen\\
    Tübingen, Germany
}
\affil[2]{%
    Technical University of Munich\\
    Munich, Germany
}
\affil[$\dagger$]{%
    equal contribution
}

\begin{document}
\maketitle

\begin{abstract}
Interest in understanding and factorizing learned embedding spaces through conceptual explanations is steadily growing. When no human concept labels are available, concept discovery methods search trained embedding spaces for interpretable concepts like \textit{object shape} or \textit{color} that can provide post-hoc explanations for decisions. 
Unlike previous work, we argue that concept discovery should be \emph{identifiable}, meaning that a number of known concepts can be provably recovered to guarantee reliability of the explanations.
As a starting point, we explicitly make the connection between concept discovery and classical methods like Principal Component Analysis and Independent Component Analysis by showing that they can recover independent concepts under non-Gaussian distributions. For dependent concepts, we propose two novel approaches that exploit functional compositionality properties of image-generating processes.
Our provably identifiable concept discovery methods substantially outperform competitors on a battery of experiments including hundreds of trained models and dependent concepts, where they exhibit up to 29\,\% better alignment with the ground truth. Our results highlight the strict conditions under which reliable concept discovery without human labels can be guaranteed and provide a formal foundation for the domain. 
Our code is available \href{https://github.com/tleemann/identifiable_concepts}{online}.
\end{abstract}

\section{Introduction}\label{sec:intro}
Modern computer vision systems represent images in embedding spaces. These are either constructed implicitly in higher-level layers of large models or explicitly through generative models such as Variational Autoencoders \citep{kingma2013auto} or recent Diffusion Models \citep{song2019generative, ho2020denoising}.
To unveil why an image is considered similar to a certain class, interest in understanding these embeddings is increasing. Conceptual explanations \citep{crabbe2022concept, muttenthaler2022vice, akula2020cocox, kazhdan2020now, yeh2019completeness, Kim2018interpretabilityTCAV} are a popular explainable AI (XAI) technique for this purpose. They scrutinize a given encoder by decomposing its embedding space into interpretable concepts post-hoc, i.e., after training. 
Subsequently, these concepts form the basis of popular post-hoc explanations such as TCAV \citep{Kim2018interpretabilityTCAV} or allow high-level interventions \citep{koh2020concept}. \cref{fig:conceptexplanation} outlines a real-world example. A misclassification made by a pretrained model shipped with the \texttt{pytorch} library \citep{paszke2017automatic} is to be explained. In the given example, the conceptual explanation allows identification of a spurious correlation that the model has picked up: Most jack-o-lanterns are found in combination with dark backgrounds, which causes it to mistake the traffic light at night for a jack-o-lantern.

\definecolor{officered}{RGB}{192,0,0}
\begin{figure*}[t]
\begin{subfigure}[b]{0.20\textwidth}
    \scalebox{0.70}{
    \begin{tikzpicture}
    \node[inner sep=0pt] (lantern) at (0,0)
    {\includegraphics[width=4.5cm]{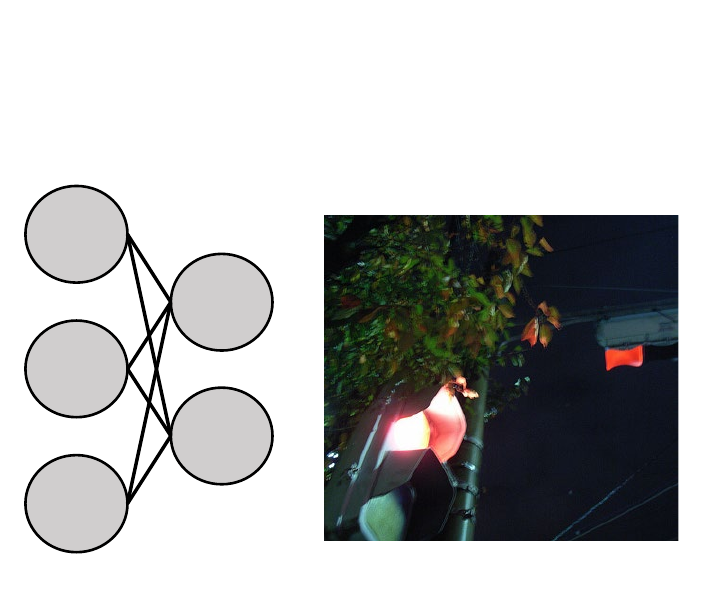}};
    \node at (0.5, 1.1) {\parbox{5cm}{\centering prediction: \textbf{jack-o-lantern}\\true class: \textbf{traffic light}}};
    \end{tikzpicture}}
    \caption{\textbf{Misclassification:} A\newline model makes an incorrect prediction. A user is interested in understanding why this incident happened.\newline~}
\end{subfigure}\hspace{2mm}
\begin{subfigure}[b]{0.25\textwidth}
    \scalebox{0.70}{
    \begin{tikzpicture}
    \node[inner sep=0pt] (lantern) at (0,0)
    {\includegraphics[width=6cm]{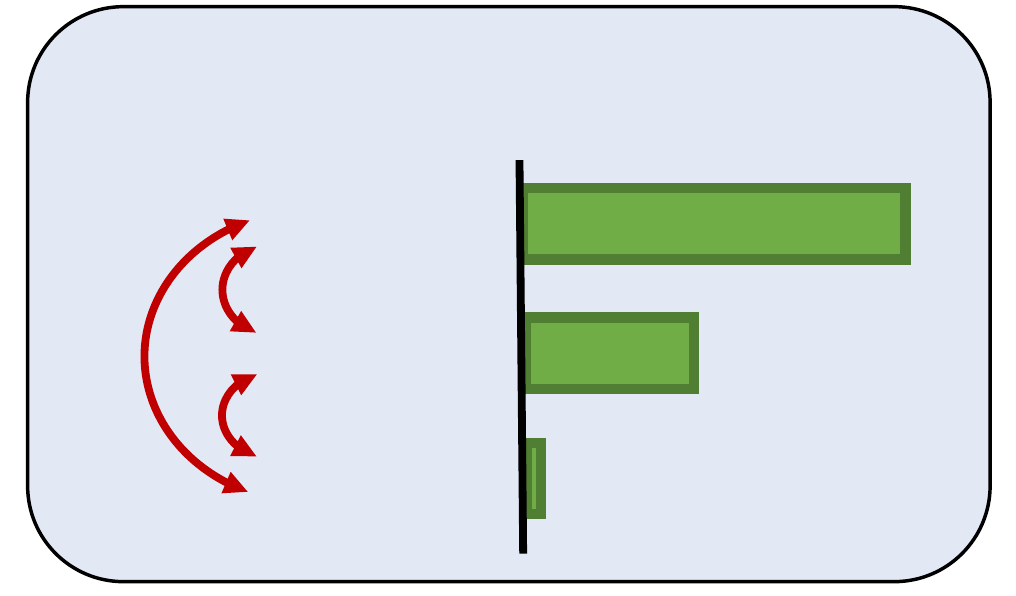}};
    \node at (0.0, 1.2) {\parbox{6cm}{\centering concept contributions for prediction\\\textbf{jack-o-lantern}}};
    \node at (-0.65, 0.48) {darkness};
    \node at (-0.6, -0.35) {fire, red};
    \node at (-0.7, -1.1) {pumpkins};
    \node[rotate=90] at (-2.4, -0.3) {\textcolor{officered}{dependencies}};
    \end{tikzpicture}}
    \caption{\textbf{Conceptual Explanation:} 
    Concept contributions are computed that explain the prediction. In this example, the concept ``darkness'' is relevant for the outcome.}
\end{subfigure}\hspace{2mm}
\begin{subfigure}[b]{0.23\textwidth}
    \centering
    \scalebox{0.70}{
    \begin{tikzpicture}
    \node[inner sep=0pt] (lantern) at (0,0)
    {\includegraphics[width=4cm]{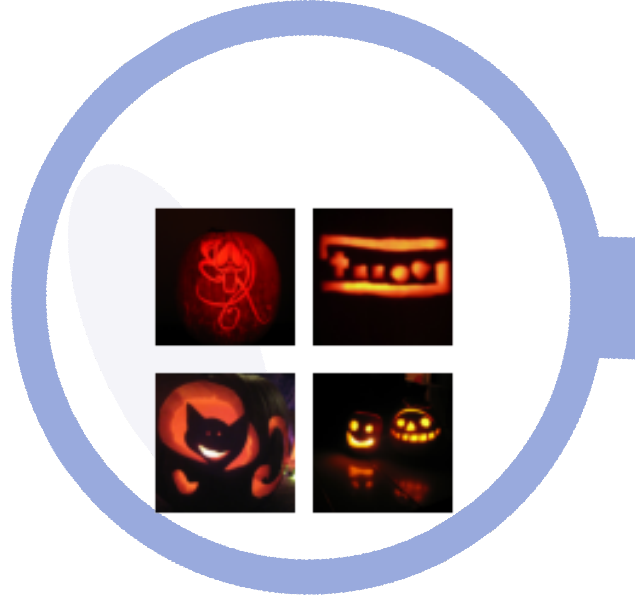}};
    \node at (-0.1, 1.05) {\parbox{5cm}{\centering examples of\\\textbf{jack-o-lantern}}};
    \end{tikzpicture}}
    \caption{\textbf{Inspection:} A closer inspection of samples from the predicted class reveals that most images in this class have a dark background; a spurious correlation picked up by the model.}
\end{subfigure}
\hspace{2mm}
\begin{subfigure}[b]{0.25\textwidth}
    \scalebox{0.70}{
    \begin{tikzpicture}
    \node[inner sep=0pt] (lantern) at (0,0)
    {\includegraphics[width=6cm]{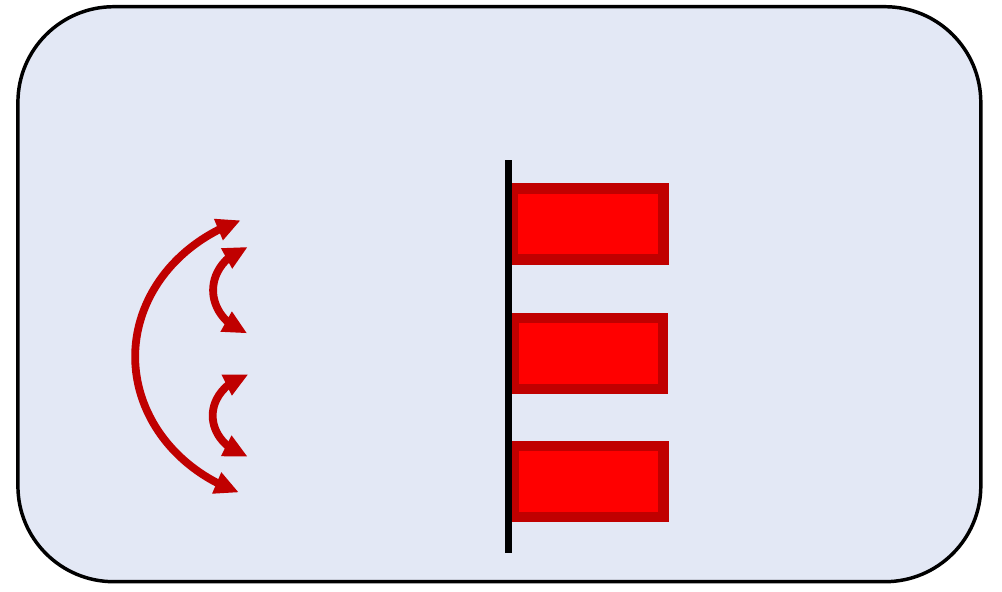}};
    \node at (0.0, 1.2) {\parbox{6cm}{\centering concept contributions for prediction\\\textbf{jack-o-lantern}}};
    \node at (-0.65, 0.48) {darkness};
    \node at (-0.6, -0.35) {fire, red};
    \node at (-0.7, -1.1) {pumpkins};
    \node[rotate=90] at (-2.4, -0.3) {\textcolor{officered}{dependencies}};
    \end{tikzpicture}}
    \caption{\textbf{Entangled Conceptual Explanation:} It is essential to correctly split up the contribution of individual concepts to allow for valid inferences.\newline\label{fig:motivationd}}
\end{subfigure}
\caption{Schematical use-case of conceptual explanations: A misclassification of an image classifier is explained. The example is based on a real explanation for a ResNet50 model. Details and the original explanation are provided in App.~\ref{sec:app_detailsintrofig}.\label{fig:conceptexplanation}}
\end{figure*}
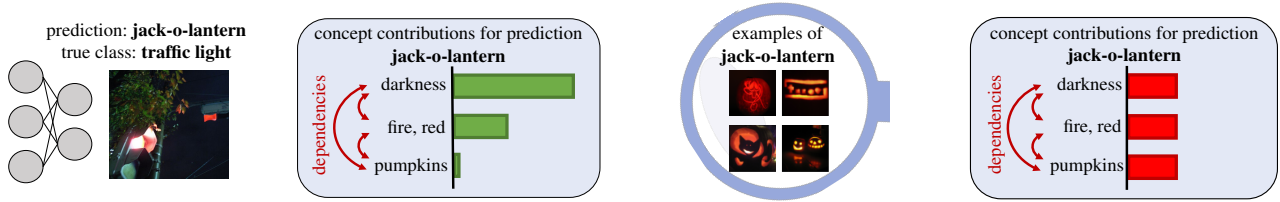

Constructing such explanations is non-trivial.
The key ingredient to all conceptual explanation techniques is a set of interpretable concepts, which is notoriously hard to specify \citep{leemann2022coherence}. It is frequently defined through human annotations \citep{crabbe2022concept, koh2020concept, Kim2018interpretabilityTCAV} on individual samples of the dataset that can be prohibitively expensive \citep{kazhdan2021disentanglement}. Furthermore, it is usually unknown which concepts will be leveraged by a machine learning model without a model at hand. Therefore, we consider fully unsupervised concept discovery \citep{ghorbani2019towards, yeh2019completeness}, where the concepts are automatically discovered in the data. 
Concepts are frequently modeled as directions in a given embedding space \citep{ghorbani2019towards,Kim2018interpretabilityTCAV, yeh2019completeness}, which have to be discovered without supervision. These embedding spaces can be highly distorted, making it hard to correctly separate the influences of individual concepts. However, this is essential to make the right inferences in practice (see \cref{fig:motivationd}). \added{This intuition is supported by prior work on generative models \citep{ross2021evaluating}, which has shown that user understanding is strongly linked to the representations' respective disentanglement.}

While many methods have been empirically shown to work well, a rigorous theoretical analysis of the conditions under which concept discovery is possible is still lacking in previous works. We propose to consider concept discovery methods that are \textit{identifiable}. This means when a known number of \textit{ground truth components} generated the data, the concept discovery method provably yields concepts that correspond to the individual ground truth components and can correctly represent an input in the concept space. This is a crucial requirement: If a method is even incapable of recovering known components, there is no indication for its reliability in practice. In this work, we are the first to investigate identifiability results in the context of post-hoc concept discovery.

First, we find that identifiability results from Principal Component Analysis (PCA) and Independent Component Analysis (ICA) literature \citep{jolliffe2002principal, comon1994independent, hyvarinen2001independent} can be transferred to the conceptual explanation setup. We establish that they cover the case of independent ground truth components with non-Gaussian distributions. This is insufficient for two reasons: (1) In practice, concepts such as height and weight \citep{trauble2021disentangled} or wing and head colors of birds often follow complex dependency patterns. (2) Popular generative models \citep{kingma2013auto, song2019generative} frequently work with an embedding space with a Gaussian distribution.

As a second contribution, we seek to fill this void by providing an identifiable concept discovery approach that can handle dependent and Gaussian ground truth components. We can show that this is possible through taking the nature of the image-generating process into consideration. Specifically, we propose utilizing \emph{visual compositionality properties}. These are based on the observation that tiny changes in the components frequently affect input images in orthogonal or even disjoint ways. These properties of image-generating processes also leave a ``trace'' in the encoders learned from a set of data samples. This insightful finding permits to construct two novel post-hoc concept discovery methods based on the \emph{disjoint} or \emph{independent mechanisms} criterion. We prove strong identifiability guarantees for recovering components, even if they are dependent. \added{Our results highlight the strict and nuanced conditions under which identifiable concept discovery is possible.}

In summary, our work advances current literature in multiple ways: 
\textbf{(1)} We present first identifiability results for post-hoc conceptual explanations. We find that results from ICA can be transferred under the assumption of independent ground truth components.
\textbf{(2)} For the more intricate setting of dependent components, we propose the \textit{disjoint mechanism analysis (DMA)} criterion and the less constrained \emph{independent mechanism analysis (IMA)} criterion. We prove that they recover even dependent original components up to permutation and scale.
\textbf{(3)} We construct DMA and IMA-based concept discovery algorithms for encoder embedding spaces with the same theoretical identifiability guarantees. 
\textbf{(4)} We test them (i) on embeddings of several autoencoder models learned from correlated data, (ii) with multiple and strong correlations, (iii) on discriminative encoders, and (iv) on the real-world CUB-200-2011 dataset \citep{Wah2011}. Our approaches maintain superior performance amidst increasingly severe challenges.

\section{Related Work}
\label{sec:relatedwork}
Works on the analysis and interpretation of embedding spaces touch a variety of subfields of machine learning.

\textbf{Concept discovery for explainable AI.} Conceptual explanations \citep{koh2020concept, Kim2018interpretabilityTCAV, ghorbani2019towards, yeh2019completeness, akula2020cocox, chen2020concept} have gained popularity within the XAI community. They aim to explain a trained machine learning model post-hoc in terms of human-friendly, high-level concept directions \citep{Kim2018interpretabilityTCAV}. These concepts are found via supervised \citep{koh2020concept, kim2018disentangling, kazhdan2020now} or unsupervised approaches \citep{yeh2019completeness, akula2020cocox, ren2022learning}, such as clustering of embeddings \citep{ghorbani2019towards}. However, their results are not always meaningful \citep{leemann2022coherence, yeh2019completeness}. Therefore, we suggest approaches with identifiability guarantees. We provide initial identifiability results and a novel approach, which can be used for unsupervised concept discovery under correlated components. 

\textbf{Independent Component Analysis (ICA).} Independent Component Analysis \citep{comon1994independent, hyvarinen1999, hyvarinen2001independent} or blind source separation (BSS) consider a generative process $\vg(\vz)$ as a mixture to undo and rely on traces that the distribution of the generating components $\vz$ leaves in the mixture.
In this work, we show that an identifiability result from ICA can be transferred to the conceptual explanation setup, but recovery is only possible under independent underlying components of which all but one are non-Gaussian. This result is not applicable to naturally correlated processes, which is why we design a novel method for this case.

\textbf{Disentanglement Learning.} Concurrently, literature on disentanglement learning is concerned with finding a data-generating mechanism $\vg(\vz)$ and a latent representation $\vz$ for a dataset, such that each of the original components (also known as factors of variation) is mapped to one (controllable) unit direction in $\vz$ \citep{bengio2013representation}. An alternative definition relies on group theory \citep{higgins2017beta} where certain group operations (symmetries) should be reflected in the learned representation \citep{painter2020LinearGroupDisentanglement, yang2021towards}. Most works in the domain enhance VAEs \citep{kingma2013auto} with additional loss terms \citep{higgins2017beta, burgess2018understanding, kim2018disentangling, chen2018isolating}.
Despite recent progress it is not always possible to construct disentangled embedding spaces from scratch: \cite{locatello2019challenging} have shown that the problem is inherently unidentifiable without additional assumptions. A more recent work by \citet{trauble2021disentangled} shows that even if just two components of a dataset are correlated, current disentanglement learning methods fail. In this work, we focus on post-hoc explanations of embedding spaces of given models, which are usually entangled.

\textbf{Identifiability results.} \added{Identifiability questions have been raised in domains such as Natural Language Processing \citep{carrington2019invariance} or in disentanglement learning, which is most related to this work.} It has been previously shown that unsupervised disentanglement, without further conditions, is impossible \citep{hyvarinen1999, locatello2019challenging, moran2022identifiable}. Hence, recent works aim to understand the conditions sufficient for identifiability. One strain of work relies on additional supervision, i.e., access to an additional observed variable \citep{hyvarinen2019nonlinear, khemakhem2020variational} or to tuples of observations that differ in only a limited number of components \citep{locatello2020weakly}. 
\citet{gresele2021independent} and \citet{zheng2022on} proved identifiable disentanglement under independently distributed components and introduce a functional condition on the data generator.
We also consider functional properties, but our setting is different as (1) we have access to a trained encoder only and (2) not even partial annotations or relations are available.

\section{Analysis}
\label{sec:theory}
In this section, we formalize post-hoc concept discovery to provide an identifiability perspective. We find that Independent Component Analysis (ICA) and Principal Component Analysis (PCA) only guarantee identifiability when the ground-truth components are stochastically independent. We then study the intricate case of dependent components and propose using \emph{disjoint} and \emph{independent mechanisms analysis} (DMA / IMA) along with identifiability results. 

\begin{figure*}[t]
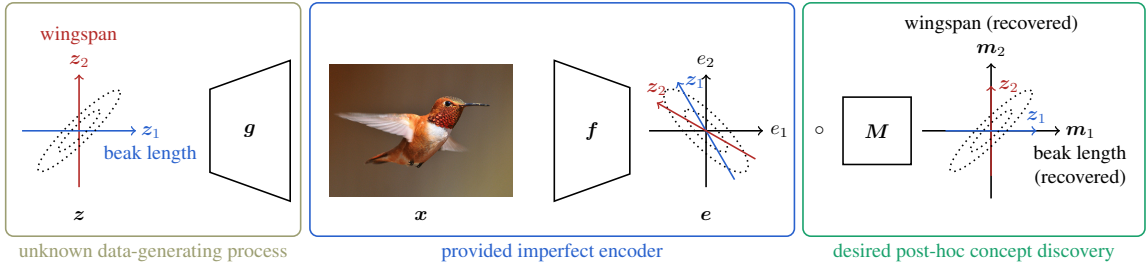

    \centering
    \ifdefined\arxiv
    \scalebox{0.75}{\input{figures/setup_figure}}
    \else
    \scalebox{0.75}{\input{figures/setup_figure}}
    \fi
    \caption{Overview over the concept discovery setup. We consider a process where data samples $\bm{x}$ are generated from possibly correlated ground truth components $\bm{z}$, e.g., a wingspan or beak length of a bird, by an unknown process $\vg$ (left). The high-dimensional data is mapped to the to the embedding space of a given model $\vf$ (center). A suitable post-hoc concept discovery yields concept vectors $\vm_i$ that correspond to the original components (right).}
    \label{fig:encdecsetup}
\end{figure*}

\subsection{Problem Formalization} \label{sec:form}

In post-hoc concept discovery, we are given a trained encoder $\vf:\mathcal{X} \rightarrow \mathcal{E}$ with embeddings $\ve = \vf(\vx) \in \mathcal{E} \subset \mathbb{R}^K$ of each image $\vx \in \mathcal{X}$. We do not impose any restriction on how $\vf$ was obtained; it can be the feature extractor part of a large classification model, or a feature representation learned through autoencoding, contrastive learning \citep{chen2020simple}, or related techniques. Interpretability literature seeks to understand the embedding space by factorizing it into concepts. Based on the observations that directions in the embedding space often correspond to meaningful features \citep{szegedy2013intriguing, bau2017network, alain2016understanding, bisazza2018lazy}, these concepts are frequently defined as direction vectors $\bm{m}_i$ \citep{Kim2018interpretabilityTCAV, ghorbani2019towards, yeh2019completeness}. \added{These are commonly referred to as concept activation vectors (CAVs).} Hence, the combined output of a concept discovery algorithm is a matrix $\mM =[\bm{m}_1, \ldots, \bm{m}_K]^\top\in \mathbb{R}^{K \times K}$ where each row contains a concept direction.

We seek a theoretical guarantee on when these discovered concept directions align with ground truth components that generated the data. To this end, we formalize the data-generating process as shown in \cref{fig:encdecsetup}: There are $K$ ground-truth components with scores $z_k, k=1\ldots K$, summarized $\vz \in \mathcal{Z} \subset \mathbb{R}^K$, that define an image. The term \textit{components} always refers to the ground truth as opposed to the \textit{concepts}, which denote the discovered directions. A data-generating process $\vg:\mathcal{Z} \rightarrow \mathcal{X}$ generates images $\vx = \vg(\vz) \in \mathcal{X} \subset \mathbb{R}^L$, $L \gg K$. A powerful algorithm should be able to recover the original components. That is, there should be a one-to-one mapping between entries of $\mM \ve$ and the entries in $\bm{z}$, up to the arbitrary scale and order of the entries. We say that a concept discovery algorithm \emph{identifies} the true components if it is guaranteed to output directions $\mM$ that satisfy $\mM \ve = \mM \vf(\vg(\vz)) = \mP\mS \vz$ $\forall \vz \in \mathcal{Z}$, where $\mP \in \mathbb{R}^{K \times K}$ is a permutation matrix that has one $1$ per row and column and is $0$ otherwise, and $\mS \in \mathbb{R}^{K \times K}$ is an invertible diagonal scaling matrix. 

To make the problem solvable in the first place, concept directions must exist in the embedding space of the given encoder, requiring $ \bm{e} = \mD \vz$, where $\mD \in \mathbb{R}^{K \times K}$ is of full rank. Depending on the scope of the conceptual explanation desired, it can be sufficient for the components to exist in a local region of the embedding space if the concept discovery algorithm is only applied around a region around a certain point of interest. This only changes the meaning of $\mathcal{E}, \mathcal{X},$ and $\mathcal{Z}$ but is formally equivalent.

\subsection{Identifiability via Independence}
Initially, we turn towards classical component analysis methods. We find that their identifiability results \added{use non-correlation or even stronger stochastic independence assumptions of the ground truth components.} 

Principal Component Analysis (PCA) \citep{jolliffe2002principal} uses eigenvector decompositions to find orthogonal directions $\mM$ that result in uncorrelated components $\mM \ve$. This means that PCA is only capable of identifying the original components if the ground truth components $\vz$ were uncorrelated and exist as orthogonal directions in our embedding space. In our setup and notation, this leads to the following result:
 
\begin{theorem}[PCA identifiability]\label{thm:ident_pca}
     Let $z_k, k=1, \dotsc, K,$ be uncorrelated random variables with non-zero and unequal variances. Let $\bm{e} = \bm{D}\bm{z}$, where $\bm{D} \in \mathbb{R}^{K \times K}$ is an orthonormal matrix. If an orthonormal post-hoc transformation $\bm{M}\in \mathbb{R}^{K \times K}$  results in mutually uncorrelated components $(z'_1, \dotsc, z'_K) = \bm{z}' = \bm{M}\bm{e}$, then $\bm{M}\ve = \bm{P}\bm{S} \vz$, where $\mP \in \mathbb{R}^{K \times K}$ is a permutation and $\bm{S} \in \mathbb{R}^{K \times K}$ is a diagonal matrix where $|s_{ii}|=1$ for $i \in 1,\ldots K$.\footnote{To simplify notation, $\mP$ and $\mS$ mean \textit{any} permutation and scale matrices. They do not have to be equal between the theorems.} 
\end{theorem}

All proofs in this work are deferred to App.~\ref{sec:app_proofs}. It is arguably a strong condition that the ground truth directions are encoded orthogonally in the embedding space. Independent Component Analysis (ICA) overcomes this limitation and allows for arbitrary directions. \added{However, the classic result by \cite{comon1994independent} even demands stochastically independent components. Transferred to our setup and notation, the result can be stated as follows.}
\begin{theorem}[ICA identifiability]\label{thm:ident_ica}
    Let $z_k, k=1, \dotsc, K,$ be independent random variables with non-zero variances where at most one component is Gaussian. Let $\bm{e} = \bm{D}\bm{z}$, where $\bm{D} \in \mathbb{R}^{K \times K}$ has full rank.  
    If a post-hoc transformation $\bm{M} \in \mathbb{R}^{N\times N}$ results in mutually independent components $(z'_1, \dotsc, z'_K) = \bm{z}' = \bm{M}\bm{e}$, then $\bm{M}\ve = \bm{P}\bm{S} \vz$, where $\mP \in \mathbb{R}^{K \times K}$ is a perm. and $\bm{S} \in \mathbb{R}^{K \times K}$ is a diag. matrix.
\end{theorem}

This result shows that stochastic independence of the ground truth components leaves a strong trace in the embeddings that can be leveraged. Algorithms like \texttt{fastICA} \citep{hyvarinen1997fast} can find the concept directions $\mM$ by searching for independence \citep{comon1994independent}. We conclude that ICA is suited for post-hoc concept discovery under independent components.

In summary, we have transferred two results from the component analysis literature to the setup of post-hoc conceptual explanations. However, these results do not allow to recover components that are correlated or follow a Gaussian distribution. This limits their applicability in practice where concepts often appear pairwise (e.g., darkness and jack-o-lanterns, cf. \cref{fig:conceptexplanation}). We will bridge this gap in the remainder of this paper by introducing two new identifiable discovery methods based on functional properties of the generation process that we term \textit{disjoint} and \textit{independent} mechanisms. A summary of identifiability results is provided in \cref{tab:summary}.
\begin{table}[t]
\adjustbox{width=\columnwidth}{
\setlength{\tabcolsep}{2pt}
\begin{tabular}{cccl}
    \toprule
    Dependency & Marginal Dist. & Transform & Criterion\\
    \midrule
    uncorr. & uneq. variances & orthogonal & non-correlation (PCA) \\
    independent & non-Gaussian & invertible & independence (ICA) \\
    arbitrary & arbitrary & invertible &  disj. mechanisms (DMA) \\
    arbitrary & arbitrary & invertible &  indep. mechanisms (IMA) \\
    \bottomrule
\end{tabular}
}
\caption{PCA and ICA provably identify concepts via their distributions. DMA and IMA utilize functional properties.}
\label{tab:summary}
\end{table}

\subsection{Identifiability via Disjoint Mechanisms}
\label{sec:transfer}
Instead of placing independence assumptions on $\vz$, we propose a concept discovery algorithm that makes use of natural properties of the generative process $\vg$. 
In particular, generative processes in vision are often compositional \citep{ommer2007learning}: Different groups of pixels in an image, like a bird's wings, legs, and head, are each controlled by different components. Effects of tiny changes in components are visible in the Jacobian $\mJ_\vg$, where each row points to the pixels affected. Thus, a compositional process will follow the \textit{disjoint mechanisms} principle. 

\begin{definition}[Disjoint mechanism analysis (DMA)] \label{def:dma}
$\vg$ is said to generate $\vx$ from its components $\vz$ via disjoint mechanisms if the Jacobian $\mJ_\vg(\vz) \in \mathbb{R}^{L\times K}$ exists and is a block matrix $\forall \vz \in \mathcal{Z}$. That is, the columns of $\mJ_\vg(\vz)$ are non-zero at disjoint rows, i.e. $| \mJ_\vg(\vz) |^\top | \mJ_\vg(\vz) | = \mS(\vz)$, where $\mS \in \mathbb{R}^{K \times K}$ is a diagonal matrix that may be different for each $\vz$ and $| \cdot |$ takes the element-wise absolute value.
\end{definition}
Note that this definition does not globally constrain the location of affected pixels. The components may still alter different but disjoint pixels for each image.
In real concept discovery, we do not have access to the generative process $\vg$ but can only access the encoder $\vf$. However, an encoder corresponding to $\vg$ will not be arbitrary and its Jacobian $\jacf \in \mathbb{R}^{K\times L}$ will have a distinct form in practice: First, to maintain the component information the composition $\vf \circ \vg$ will be of the form $\vf(\vg(\vz))=\bm{D}\bm{z}$, with a yet unknown matrix $\bm{D} \in \mathbb{R}^{K\times K}$. Furthermore, we expect encoders to be rather lazy, meaning they only perform the changes to invert the data generation process but are almost invariant to input deviations not due to changes in the components. \added{This is in line with the classic interpretability literature, where gradients of models were observed to noisily highlight the relevant input features \citep{baehrens2010explain, simonyan2013deep} and form the basis of popular attribution methods such as Integrated Gradients \citep{Sundararajan2017}.} Technically, the changes effected by the components form the linear $\text{span}(\mJ_\vg(\vz))$, whereas entirely external changes are given in its orthogonal complement $\text{span}(\mJ_\vg(\vz))^{\perp}$. Thus, for $\mathbf{v} \in \text{span}(\mJ_\vg(\vz))^\perp \subset \mathbb{R}^L$ the encoder should not react to these change and \added{the corresponding gradients of the encoder for these changes should be zero, i.e., $\mJ_\vf(\vg(\vz))\mathbf{v} = \mathbf{0} \Leftrightarrow \mathbf{v} \in \text{ker}(\mJ_\vf(\vg(\vz)))$.}
\begin{definition}[Faithful encoder]
    $\vf$ is a faithful encoder for the generative process $\vg$ if the ground truth components remain recoverable, i.e., $\vf(\vg(\vz))=\bm{D}\bm{z}$, for some $\bm{D} \in \mathbb{R}^{K \times K}$ with full rank. Furthermore, $\bm{f}$ is lazy and invariant to changes in $\bm{x}$ which cannot be explained by the ground truth components, requiring $\mJ_{\vf}(\vg(\vz))$ and $\mJ_\vg(\vz)$ to exist and $ \text{\normalfont span}(\mJ_\vg(\vz))^\perp \subseteq \text{\normalfont ker}(\mJ_\vf(\vg(\vz))),~ \forall \vz \in \mathcal{Z}$.
\end{definition}

\added{Having defined what realistic encoders look like through the notion of faithful encoders,} we find that there is distinct property which can be leveraged to discover the directions in $\mM$ among faithful encoders: \added{It is sufficient to find an encoder $\mM\vf$ whose Jacobian $\mM\mJ_{f}$ will have disjoint rows. Intuitively, this means searching for components whose gradients affect disjoint image regions.}

\begin{theorem}[Identifiability under DMA] \label{eq:ident_full}
Let $\vg$ have disjoint mechanisms and $\vf$ be a faithful encoder to $\vg$. 
If a post-hoc transformation $\bm{M} \in \mathbb{R}^{K\times K}$ of full rank results in disjoint rows in the Jacobian $\mM \mJ_{\vf}(\vg(\vz))$, i.e., $\lvert\mM \mJ_{\vf}(\vg(\vz))\rvert\lvert\mM \mJ_{\vf}(\vg(\vz))\rvert^\top$ is invertible and diagonal for some $\vz \in \mathcal{Z}$, then $\bm{M}\ve = \bm{P}\bm{S} \vz$ where $\mP \in \mathbb{R}^{K \times K}$ is a permutation and $\bm{S} \in \mathbb{R}^{K \times K}$ is a scaling matrix.
\end{theorem}

This theorem does not impose any restrictions on the distribution $\vz$, making it applicable to realistic concept discovery scenarios through leveraging the nature of the generative process. The proof of this algorithm in App.~\ref{sec:app_proofdma} also yields an analytical solution. We will use it to verify conditions in a controlled experiment in \cref{sec:synthdata}. We have thus identified the \textit{DMA criterion} that \added{is sufficient} to discover the component directions when the rows of $\mM \mJ_\vf$ point to disjoint image regions. We can formulate this as a loss function and optimize for $\mM$ via off-the-shelf gradient descent: 
\begin{align}\label{eqn:objective}
    \mathcal{L}(\mM) = \mathbb{E}_\vx\lVert \text{arn}\left[\mM \jacf(\vx)\right]\text{arn}\left[\mM \jacf(\vx)\right]^\top - \mI \rVert_F^2.
\end{align}
The expectation is taken over a collection of real data samples $\vx=\vg(\vz)$. The $\text{arn}$-operator (\underline{a}bsoute values, \underline{r}ow \underline{n}ormalization) takes the element-wise absolute value and subsequently normalizes the rows. This does not constrain the norms of the Jacobian's rows but only enforces disjointness.

\subsection{Concept Discovery via Independent Mechanisms} \label{sec:theoryOrtho}
We can perform an analogous derivation for a class of generating processes that is more general. Grounded by causal principles instead of compositionality, the independent mechanisms property has been argued to define a class of natural generators \citep{gresele2021independent}.

\begin{definition}[Independent mechanism analysis (IMA)]
\label{def:ima}
$\vg$ is said to generate $\vx$ from its components $\vz$ via independent mechanisms if the Jacobian $\mJ_\vg(\vz)$ of $\vg$ exists and its columns (one per component) are orthogonal $\forall \vz \in \mathcal{Z}$, i.e., $\mJ_\vg^\top(\vz) \mJ_\vg(\vz) = \mS(\vz)$, where $\mS \in \mathbb{R}^{K \times K}$ is a diagonal matrix that may differ for each $\vz$ \citep{gresele2021independent}.
\end{definition}

\citet{gresele2021independent} and \citet{zheng2022on} used this characteristic to find disentangled data generators, but we can again transfer characteristics via faithful encoders: This time we find that searching for an $\mM\mJ_{f}$ with \textit{orthogonal} (instead of disjoint) rows permits post-hoc discovery of concepts. We refer to is property of $\mM\mJ_{f}$ as the \textit{IMA criterion}. 

However, as the class of admissible processes has been increased, it is not strong enough to ensure identifiability in the most general case. This is prevented under an additional technical condition on the component magnitudes, which we refer to as \emph{non-equal magnitude ratios} (NEMR). Intuitively, the magnitudes of the component gradients have to change non-uniformly between at least two points \added{for the conditions to be sufficient}. \added{If there were two factors that always attribute to input pixels in the same way (imagine the sky being partitioned into two components termed ``left sky'' and ``right sky''), they cannot be told apart anymore since there can be other mixtures which would result in orthogonality (they could equally be ``lower sky'' and ``upper sky'').}

\begin{theorem}[Identifiability under IMA] \label{thm:ident_ima}
Let $\vg$ adhere to IMA. Let $\vf$ be a faithful encoder to $\vg$. 
Suppose we have obtained an $\vf'= \mM \vf$ with a full-rank $\mM \in \mathbb{R}^{K \times K}$ and orthogonal rows in its Jacobian  $\mM \mJ_{\vf}(\vg(\vz)) \coloneqq \jacfprime(\vg(\vz))$, i.e,  $\jacfprime(\bm{g}(\bm{z}))\jacfprime(\bm{g}(\bm{z}))^\top = \bm{\Sigma}(\bm{z})$ where $\bm{\Sigma}(\bm{z})$ is diagonal and full-rank at two points $\vz \in \{\vz_a, \vz_b\}$. If additionally $\bm{\Sigma}(\bm{z}_a) \bm{\Sigma}(\bm{z}_b)^{-1}$
has unequal entries in its diagonal (NEMR condition), then $\bm{M}\ve = \bm{P}\bm{S} \vz$, where $\mP \in \mathbb{R}^{K \times K}$ is a permutation and $\bm{S} \in \mathbb{R}^{K \times K}$ is a scaling matrix.
\end{theorem}

The constructive proof in App.~\ref{sec:app_proofima} can also be condensed into an analytical solution. Alternatively, one can again construct a suitable optimization objective for the IMA criterion, i.e., orthogonal Jacobians. This is achieved by removing the absolute value operation from the arn-operator in \cref{eqn:objective}, so that it solely performs a row-wise normalization. In summary, we have established the novel DMA and IMA criteria that allow concept discovery under dependent components.

\section{Experiments}
In the following, we perform a battery of experiments of increasing complexity to compare the practical capabilities of approaches for identifiable concept discovery. We start by verifying the theoretical identifiability conditions (\cref{sec:synthdata}), then perform evaluation under increasing multi-component correlations for embedding spaces of generative and discriminative models (\cref{sec:exp_comp} to \ref{sec:exp_discr}), and finally use a large-scale, discriminatively-trained ResNet50 encoder (\cref{sec:exp_cub}). 

We borrow the DCI metric \citep{eastwood2018framework} from disentanglement learning 
with scores in $[0,1]$ to measure whether each discovered component predicts precisely one ground-truth component and vice versa. Following \citet{locatello2020weakly}, 
we report additional metrics with similar results in App.~\ref{sec:app_additionalresults}, along with results on additional datasets and ablations. For reproducibility, each experiment is repeated on five seeds and code is made available upon acceptance. In total, we train and analyze over 300 embedding spaces, requiring about 124 Nvidia RTX2080Ti GPU days. More implementation details are in App.~\ref{sec:app_expdetails}.

\newcommand{\figthreew}{1.0}
\begin{figure}[t]
    \begin{subfigure}[b]{\figthreew\linewidth}
    \begin{minipage}[t]{0.5\textwidth}
    \centering \small Traversals along each component
    \includegraphics[width=0.9\textwidth, trim=0 0.56cm 0 0, clip]{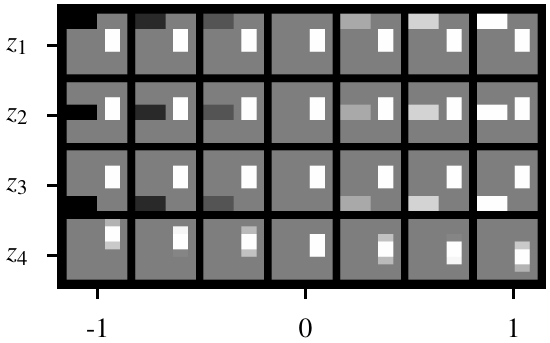}
    \end{minipage}
    \begin{minipage}[t]{0.3\textwidth}
    \small \,\,\,\,\,\,\,\,Gradients
    \includegraphics[width=\textwidth]{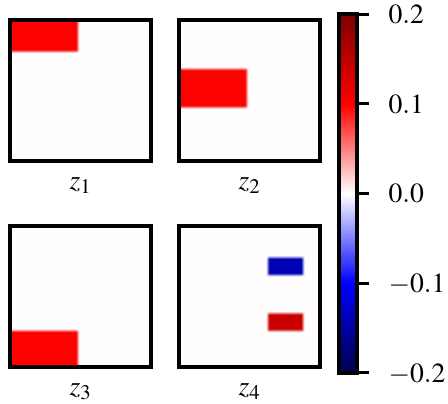}
    \end{minipage}
    \begin{minipage}[t]{0.16\textwidth}
    ~\newline
    \scalebox{0.7}{
    \begin{tabular}{c}
    DCI Scores\\
    \toprule
    IMA\\
    0.24 $\pm$ 0.10 \\
    \midrule
    DMA \\
    \textbf{1.00} $\pm$ 0.00 \\
    \bottomrule
    \end{tabular}
    }
    \end{minipage}
    \caption{\texttt{FourBars}:\,DMA datasets\,can\,be\,solved\,by\,the\,DMA\,criterion.\label{fig:fourbars}\newline~} 
    \end{subfigure}
   \begin{subfigure}[b]{\figthreew\linewidth}
    \begin{minipage}[t]{0.5\textwidth}
    \centering
    \small Traversals along each component
    \includegraphics[width=0.9\textwidth, trim=0 0.56cm 0 0, clip]{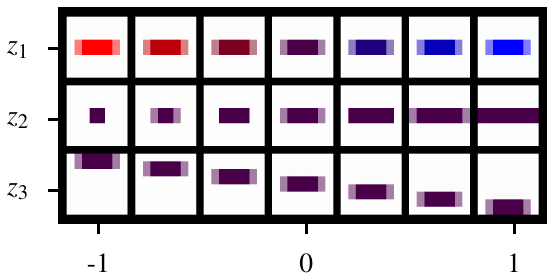}
    \end{minipage}
    \hfill
    \begin{minipage}[t]{0.3\textwidth}
    \small \,\,\,\,\,\,\,\,Gradients
    \includegraphics[width=\textwidth]{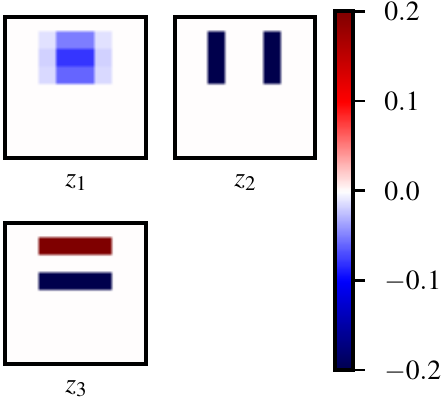}
    \end{minipage}
    \begin{minipage}[t]{0.18\textwidth}
    ~\newline
    \scalebox{0.7}{
    \begin{tabular}{c}
    DCI Scores\\
    \toprule
    IMA \\
    \textbf{1.00} $\pm$ 0.00 \\
    \midrule
    DMA\\
    0.26 $\pm$ 0.05 \\
    \bottomrule
    \end{tabular}
    }
    \end{minipage}
    \caption{\texttt{ColorBar}: IMA datasets can be solved by the IMA criterion.\label{fig:colorbar}}
    \end{subfigure}
    \caption{Experiments on two synthetic datasets: We confirm our analytical results and show that DMA (a) and IMA (b) cover visual concepts such as colors and translations.}
    \label{fig:syntheticdata}
\end{figure}

\subsection{Confirming Identifiability}\label{sec:synthdata}
We first confirm our identifiability guarantees with the analytical solutions. To this end, we implement two realistic synthetic datasets with differentiable generators. This allows computing the closed form of $\mJ_\vg$ and deliberately fulfilling or violating the DMA, IMA, and NEMR conditions. 

\texttt{FourBars} consists of gray-scale images of four components: Three bars change their colors (black to white) and one bar moves vertically, showing that the image regions affected by each component may change in each image. The plot of $\mJ_\vg$ in \cref{fig:fourbars} shows that each component maps to a disjoint image region. This fulfills DMA and thus also IMA. However, all factors have the same gradient magnitudes, making it impossible to find two points with NEMR. According to our theory, we expect DMA optimization to work and IMA to fail \added{as NEMR is essential to make to proof of \cref{thm:ident_ima}}. The second dataset, \texttt{ColorBar}, contains a single bar that undergoes realistic changes in color, width, and its vertical position, see \cref{fig:colorbar}. It conforms to IMA and NEMR but not DMA. Our proofs indicate that IMA should work, and DMA should fail. Completing the problem formalization in \cref{sec:form}, we compute analytical faithful encoders $\vf$ for these datasets distorted by a random matrix $\mD$. The solutions behave as expected: On \texttt{FourBars} only the DMA criterion delivers perfectly recovered components (DCI=1) whereas on \texttt{ColorBars} only IMA succeeds.

\subsection{Correlated Components} \label{sec:exp_comp}
We now move to the common Shapes3D \citep{3dshapes18} dataset. It shows geometric bodies that vary in their colors, shape, orientation, size, and background totaling six components. Compared to the previous section we train real encoders. We start our analysis where disentanglement learning is no longer possible: When components are correlated.
Following \citet{trauble2021disentangled}, the dataset is resampled such that two components $z_i, z_j \in [0, 1]$ follow $z_i - z_j \sim \mathcal{N}(0, s^2)$. Lower $s$ results in a stronger correlation where only few pairs of component values co-occur frequently. We choose a moderate correlation of $s=0.4$ here and three pairs $z_i, z_j$ that are nominal/nominal, nominal/ordinal, and ordinal/ordinal variables. 
We train four state-of-the-art disentanglement learning VAEs (BetaVAE \citep{higgins2017beta}, FactorVAE \citep{kim2018disentangling}, BetaTCVAE \citep{chen2018isolating}, DipVAE \citep{kumar2018variational}) from a recent study \citep{locatello2019challenging} and apply ICA, PCA, and our DMA and IMA discovery methods on their embedding spaces to post-hoc recover the original components. For DMA and IMA, we use the optimization-based algorithms (Eqn.~\ref{eqn:objective}) since they find approximate solutions through aggregation of many noisy sample gradients. 

\newcommand{\res}[2]{$#1 \pm \small#2$}
\newcommand{\posimp}[1]{\textcolor{ForestGreen}{#1}}
\newcommand{\bposimp}[1]{\textbf{\textcolor{ForestGreen}{#1}}}
\newcommand{\negimp}[1]{\textcolor{BrickRed}{#1}}
\newcommand{\bres}[2]{$\bm{#1} \pm \bm{#2}$}
\newcommand{\ures}[2]{\underline{$#1 \pm #2$}}
\newcommand{\wrapb}[2]{\begin{tabular}[c]{@{}c@{}}#1 \\#2 \end{tabular}}

\begin{table}[t]
\centering
\adjustbox{width=\figthreew\columnwidth}{
\setlength{\tabcolsep}{2pt}
\begin{tabular}{r*{6}{c}}
\toprule
\wrapb{Correlated}{components} & \multicolumn{2}{c}{\wrapb{floor \&}{background}}  &\multicolumn{2}{c}{\wrapb{orientation \&}{background}} &  \multicolumn{2}{c}{\wrapb{orientation \&}{size}}\\
\cmidrule{1-1}\cmidrule(lr){2-3}\cmidrule(lr){4-5} \cmidrule(lr){6-7}
\textbf{BetaVAE} &\res{0.497}{0.03}& & \res{0.581}{0.04}& & \res{0.491}{0.05}& \\
+PCA&\res{0.263}{0.03}&\negimp{-47\%}& \res{0.310}{0.02}&\negimp{-47\%}& \res{0.324}{0.04}&\negimp{-34\%}\\
+ICA&\res{0.574}{0.04}&\posimp{+16\%}& \res{0.540}{0.08}&\negimp{-7\%}& \res{0.577}{0.04}&\posimp{+17\%}\\
+Ours (IMA)&\res{0.617}{0.02}&\posimp{+24\%}& \res{0.602}{0.05}&\posimp{+3\%}& \res{0.579}{0.03}&\posimp{+18\%}\\
+Ours (DMA)&\bres{0.641}{0.03}&\bposimp{+29\%}& \bres{0.624}{0.06}&\bposimp{+7\%}& \bres{0.627}{0.03}&\bposimp{+28\%}\\
\cmidrule{1-1}\cmidrule(lr){2-3}\cmidrule(lr){4-5} \cmidrule(lr){6-7}
\textbf{FactorVAE} &\res{0.507}{0.11}& & \res{0.502}{0.08}& & \bres{0.712}{0.01}& \\
+PCA&\res{0.358}{0.07}&\negimp{-29\%}& \res{0.474}{0.05}&\negimp{-5\%}& \res{0.556}{0.03}&\negimp{-22\%}\\
+ICA&\res{0.294}{0.07}&\negimp{-42\%}& \res{0.263}{0.05}&\negimp{-48\%}& \res{0.340}{0.03}&\negimp{-52\%}\\
+Ours (IMA)&\res{0.551}{0.04}&\posimp{+9\%}& \res{0.498}{0.03}&\negimp{-1\%}& \res{0.595}{0.05}&\negimp{-16\%}\\
+Ours (DMA)&\bres{0.584}{0.05}&\bposimp{+15\%}& \bres{0.510}{0.05}&\bposimp{+2\%}& \res{0.556}{0.04}&\negimp{-22\%}\\
\cmidrule{1-1}\cmidrule(lr){2-3}\cmidrule(lr){4-5} \cmidrule(lr){6-7}
\textbf{BetaTCVAE} &\res{0.619}{0.01}& & \res{0.613}{0.04}& & \res{0.659}{0.01}& \\
+PCA&\res{0.400}{0.03}&\negimp{-35\%}& \res{0.421}{0.07}&\negimp{-31\%}& \res{0.450}{0.07}&\negimp{-32\%}\\
+ICA&\res{0.540}{0.02}&\negimp{-13\%}& \res{0.497}{0.04}&\negimp{-19\%}& \res{0.627}{0.02}&\negimp{-5\%}\\
+Ours (IMA)&\res{0.623}{0.02}&\posimp{+1\%}& \res{0.652}{0.03}&\posimp{+6\%}& \res{0.638}{0.04}&\negimp{-3\%}\\
+Ours (DMA)&\bres{0.666}{0.01}&\bposimp{+8\%}& \bres{0.664}{0.02}&\bposimp{+8\%}& \bres{0.748}{0.03}&\bposimp{+14\%}\\
\cmidrule{1-1}\cmidrule(lr){2-3}\cmidrule(lr){4-5} \cmidrule(lr){6-7}
\textbf{DipVAE} &\res{0.631}{0.02}& & \res{0.652}{0.02}& & \res{0.548}{0.04}& \\
+PCA&\res{0.158}{0.01}&\negimp{-75\%}& \res{0.160}{0.02}&\negimp{-75\%}& \res{0.170}{0.02}&\negimp{-69\%}\\
+ICA&\res{0.630}{0.02}&\negimp{-0\%}& \res{0.651}{0.02}&\negimp{-0\%}& \res{0.542}{0.03}&\negimp{-1\%}\\
+Ours (IMA)&\res{0.644}{0.02}&\posimp{+2\%}& \res{0.624}{0.01}&\negimp{-4\%}& \res{0.558}{0.05}&\posimp{+2\%}\\
+Ours (DMA)&\bres{0.684}{0.01}&\bposimp{+8\%}& \bres{0.679}{0.01}&\bposimp{+4\%}& \bres{0.601}{0.05}&\bposimp{+10\%}\\
\bottomrule
\end{tabular}
}
\ifdefined\arxiv
\fi

\caption{DMA recovers the components best in 11 out of 12 cases across different models and correlated components of Shapes3D. Mean $\pm$ std. err. of DCI across all components.\label{tab:posthocdisentangle}}
\end{table}

\definecolor{mygrey}{HTML}{797979}
\definecolor{mygreen}{HTML}{69cd64}
\definecolor{myorange}{HTML}{ef8649}
\definecolor{myblue}{HTML}{4878d0}
\definecolor{myred}{HTML}{ae0031}
\newcommand{\includetraversalsmall}[1]{\includegraphics[width=0.9\textwidth]{#1}}
\begin{figure}[t]
\centering
\begin{subfigure}[b]{0.48\linewidth}
\centering
\includetraversalsmall{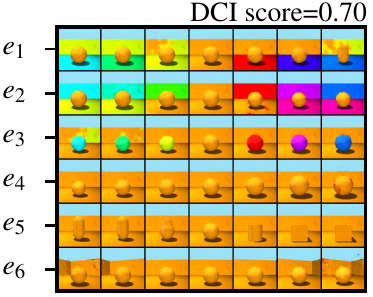}
\caption{Autoencoder (DipVAE) \label{fig:travunit}}
\end{subfigure}
\begin{subfigure}[b]{0.48\linewidth}
\centering
\includegraphics[width=0.92\textwidth]{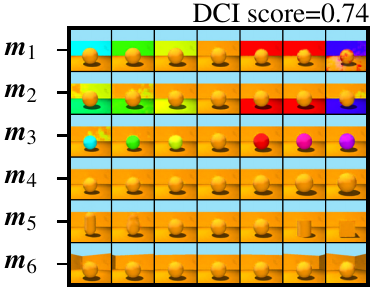}
\caption{Autoencoder + DMA\,(ours) \label{fig:travours}}
\end{subfigure}
\caption{DMA discovers directions $\vm$ that control individual concepts (wall \& floor color) of Shapes3D although they are confused in the original embedding space ($e_1$, $e_2, \ldots$).}
\label{fig:travsersal}
\end{figure}

\begin{figure}
\begin{subfigure}[b]{0.54\linewidth}
\centering
\includegraphics[height=2.35cm, trim=0 0.55cm 0 0, clip]{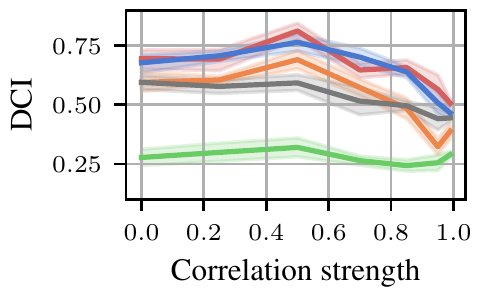}
\caption{Correlation strength} \label{fig:robustness_strength}
\end{subfigure}
\begin{subfigure}[b]{0.44\linewidth}
\centering
\includegraphics[trim={1.265cm 0.55cm 0 0},clip,height=2.35cm]{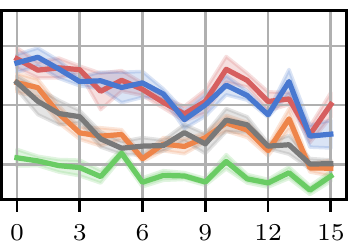}
\caption{Number pairwise corr.} \label{fig:robustness_num}
\end{subfigure}
\caption{\textcolor{myred}{DMA} and \textcolor{myblue}{IMA} recover the components even under strong and multiple correlations between them. \textcolor{myorange}{ICA} and \textcolor{mygreen}{PCA} fail to return better components than the \textcolor{mygrey}{unit axes}.} 
\end{figure}

\cref{tab:posthocdisentangle} shows the resulting DCI scores. In line with \citet{trauble2021disentangled}, we find that the disentanglement learning VAEs fail to recover the correlated components on their own due to their violated stochastic independence assumption (\cref{fig:travunit}).
In eleven of the twelve model/correlation pairs, DMA or IMA identify better concepts than the VAE unit axes and the PCA/ICA components with improvements of up to 29\,\%. This experiment shows that their concept discovery works regardless of (1) the model type and (2) the type of components correlated. On average, DMA delivers better results than IMA ($+0.047$), despite the generative process of Shapes3D only being roughly IMA or DMA-compliant. \added{We therefore hypothesize that the DMA criterion might be more robustly optimizable in practice.} \cref{fig:travours} visualizes the performance achieved via DMA when traversing the embedding space. It also shows that small DCI differences can mean a significant improvement. This is because (1) the metric is computed across all six components and the strong baselines already identify many concepts and (2) a perfect score of 1.0 is usually not possible due to non-linearly encoded components. We investigate other correlation strengths with similar findings in App.~\ref{sec:app_morerectify}.


\begin{table}[t]
     \centering
     \adjustbox{width=\figthreew\linewidth}{
     \begin{tabular}{r*{5}{c}}
     \toprule
Method & $s=0.1$ & $s=0.15$ & $s=0.2$ & $s=\infty$\\
\midrule
unit dirs.&\res{0.238}{0.01} & \res{0.244}{0.01} & \res{0.247}{0.01} & \res{0.286}{0.02}\\
PCA&\res{0.238}{0.01} & \res{0.376}{0.03} & \res{0.373}{0.03} & \res{0.343}{0.03}\\
ICA&\res{0.409}{0.02} & \res{0.309}{0.02} & \res{0.311}{0.01} & \bres{0.652}{0.00}\\
(Ours) IMA &\res{0.295}{0.01} & \res{0.302}{0.01} & \res{0.333}{0.04} & \res{0.266}{0.12}\\
(Ours) DMA &\bres{0.435}{0.01} & \bres{0.411}{0.03} & \bres{0.392}{0.02} & \res{0.369}{0.05} \\
\bottomrule
 \end{tabular}}
\caption{Without correlations ($s = \infty$), ICA is able to recover the components of a classification model. Under correlations, DMA works best. Mean $\pm$ std. err. of DCI. \label{tab:posthocdisc}}
\end{table}

\subsection{Gaussianity and Multiple Correlations} \label{sec:exp_robust}
In this section, we increase the distributional challenges to analyze whether our approaches are as distribution-agnostic as intended. We sample the components of Shapes3D from a (rotationally symmetric) Gaussian. Additionally, we introduce correlations between multiple components to its covariance matrix. Details on how covariance matrices are constructed are given in App.~\ref{sec:app_corrsampling}.  

First, we study a single pair of correlated components (floor and background color) with increasing correlation strength $\rho$. 
\cref{fig:robustness_strength} shows that the BetaVAE handles low correlations well but starts deteriorating from a strength of $\rho > 0.5$, along with ICA. The DCI of our methods is an average constant of $+0.145$ above the BetaVAE's for $\rho \leq 0.85$. After this, it returns to the underlying BetaVAE's DCI, possibly because the two components collapsed in the BetaVAE's embedding space. 
For \cref{fig:robustness_num}, we gradually add more moderately correlated ($\rho \approx 0.7$) pairs to the Gaussian's covariance matrix until eventually all components are correlated. Again, our models show a constant benefit over the underlying BetaVAE's DCI curve. This experiment highlights that both DMA and IMA perform well with (1) strong and (2) multiple correlations and (3) Gaussian components.

\begin{figure*}[t]
    \centering
    \input{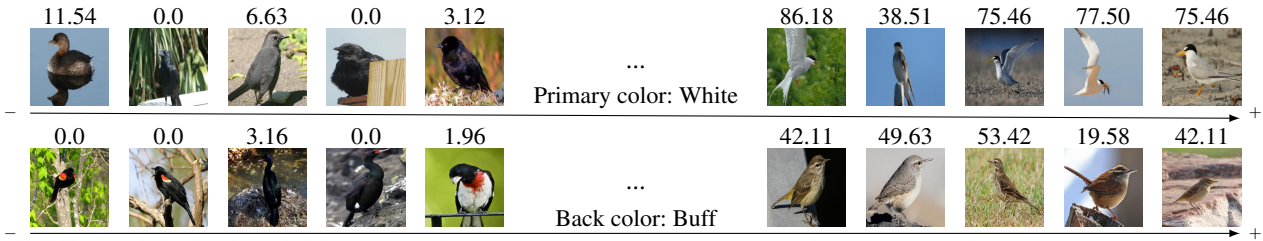}
    \caption{Components discovered by DMA on CUB correlate with interpretable ground truth attributes. Images are ordered by their concept scores $(\mM\ve)_i$, and the numbers show their ground truth annotated attribute score.}
    \label{fig:cub_attr}
\end{figure*}
\subsection{Discriminative Embedding Spaces}
\label{sec:exp_discr}

We highlight that our approach is also applicable to classification models that were trained in a purely discriminative manner, e.g., the feature space of a CNN model. 
To investigate this setting, we set up an 8-class classification problem on the Shapes3D dataset, where the combination of the four binarized components object color, wall color (blue/red vs. yellow/green), shape (cylinder vs. cube) and orientation (left vs. right) determines the class as visualized in App.~\ref{sec:app_discrsetup}. 
To make the setting even more realistic, we artificially add labeling noise close to the decision boundary, correlations as in \cref{sec:exp_comp}, and a small L2-regularizer on the embeddings to keep them in a reasonable range. We train a discriminative CNN with a $K{=}6$-dimensional embedding space.

The discriminative loss leads to a clustered distribution in the embedding space. ICA expectedly works very well in this highly non-Gaussian distribution, when no significant correlations are present which is in line with the result in \cref{thm:ident_ica}. However, tables turn as we increasingly correlate the floor and background color: Starting at $s=0.2$, DMA outperforms ICA and the other methods as can be seen in \cref{tab:posthocdisc}. While IMA leads to better concepts over the unit directions, it does not reach the level of DMA. \added{We note that both ICA and our methods improve again for very strong correlations, where the setup approaches the case of three independent components (the other two components being treated as one) that is easier again.} Overall, this demonstrates that our methods are applicable to purely discriminative embedding spaces and are more robust to high levels of correlations than ICA.

\subsection{Real-world Concept Discovery} \label{sec:exp_cub}
Last, we go beyond the traditional benchmarks and perform realistic concept discovery: We analyze the embedding space of a ResNet50 classifier \citep{He2016} trained on the CUB-200-2011 \citep{Wah2011} dataset consisting of high-resolution images of birds. This amplifies the challenges of the previous sections, i.e., a discriminative space, non-linear component dependencies of varying strengths across multiple components, and a large 512-dimensional embedding space. One restriction of this experiment is that CUB has no data-generating components to compare against, so we cannot report DCI scores. However, we qualitatively show that DMA can deliver interpretable concepts by matching them to annotated attributes of CUB.

We apply DMA and IMA to discover $K{=}30$ concepts of which the first two \added{DMA} concepts are shown exemplarily in \cref{fig:cub_attr}. The images with the highest positive scores on the first component (on the right) consistently show white birds. The other end of the component comprises birds whose primary color is black. This gives a high Spearman rank correlation with the CUB attribute ``primary color: white''. The second concept is similarly interpretable. To quantify this across all $K$ components, we provide an initial quantitative evaluation based on the Spearman rank correlation between components and attributes in App.~\ref{sec:app_cub_eval}. It indicates that ICA and PCA have problems providing such components and the components identified by DMA usually correspond more closely to the attributes. \added{The concepts provided by our method also compare favorably to those identified by ACE \citep{ghorbani2019towards} and ConceptSHAP \citep{yeh2019completeness}.} While the construction of further quantitative evaluation schemes goes beyond the scope of this work, these promising results highlight that DMA also works for high-dimensional, real-world datasets. 

\section{Discussion}

\added{We conclude by discussing the limitations of this work and related approaches and provide constructive guidance on which approach to choose in practice.}

\textbf{Limitations.} \added{In order to overcome distributional assumptions, our approach requires other forms of constraints. Most notably, we suppose that the generative processes comply with the functional properties of Disjoint or Independent Mechanisms. While they are intuitive and our empirical results suggest that they are a useful approximation of real-world images, we acknowledge that these requirements are not strictly fulfilled in most practical scenarios and the quality of the results depends on the extent to which these constraints are violated. We investigate the robustness of our methods to violations of the assumptions in App.~\ref{sec:app_ablations}. Compared to the classical methods such as PCA or IMA, the gradient-based optimization requires additional resources. However, the runtime strongly depends on hyperparameters such as the number of optimization steps. We also show that improved results can still be obtained time budget comparable to that of PCA and IMA in App.~\ref{sec:app_ablations}.}

\textbf{Choosing the right approach for concept discovery.} \added{Overall, our results show that unsupervised conceptual explanations with guarantees are only possible under specific sets of working assumptions. In this paragraph, we would like to briefly summarize them and give constructive suggestions on which approaches are best used when.
\begin{itemize}
    \item PCA works with uncorrelated components that are orthogonally encoded. We believe that the assumption of an orthogonal encoding is rather unlikely in practice, even if non-correlation was possible.
    \item ICA works well under independently distributed components but fails under dependent components. We suggest using this method when there is evidence that the ground truth components are independent.
    \item DMA does not require independence, but instead requires a disjoint mechanisms process and a faithful encoder to this process. This assumption is particularly suitable for image-generating processes.
    \item IMA does not require independence as well, but requires a faithful encoder to an independent mechanisms process. The class of independent mechanism processes is larger and may also cover non-image processes \citep{gresele2021independent}. However, it requires the additional NEMR condition. We further empirically observed that the objective derived from IMA is harder to optimize for with SGD optimizers.
    \item Other approaches like ConceptSHAP \citep{yeh2019completeness} and ACE \citep{ghorbani2019towards} also come with certain restrictions: ACE requires a model that is scale and shift invariant, while ConceptSHAP is specifically designed for computer vision models with spatial feature maps such as ResNet \citep{He2016}. Further, these approaches come without formal guarantees.
\end{itemize}
}

\section{Conclusion}
\textbf{Summary.} We proposed identifiability as a minimal requirement for concept discovery algorithms. Furthermore, we suggested the two functional paradigms of disjoint and independent mechanisms and proved that they can recover known components in visual embedding spaces. Extensive experiments confirmed that they offer substantial improvements on various generative and discriminative models and remain unaffected by distributional challenges.

\textbf{Outlook.} We believe our work to be a valuable step towards a rigorous formalization of concept discovery. However, the considered setup can be generalized in the future, for instance to components that are not linearly encoded. This would permit even stronger guarantees. While we have taken a technical perspective here, future work is required to investigate the effect of improved concepts on upstream explanations.

\begin{acknowledgements} 
The authors thank Frederik Träuble, Luigi Gresele, and Julius von Kügelgen for insightful discussions during early development of this project. This work was funded by the Deutsche Forschungsgemeinschaft (DFG, German Research Foundation) under Germany’s Excellence Strategy – EXC number 2064/1 – Project number 390727645. We thank the International Max Planck Research School for Intelligent Systems (IMPRS-IS) for supporting Michael Kirchhof.
\end{acknowledgements}

\bibliography{references}
\title{When are Post-hoc Conceptual Explanations Identifiable? (Supplementary material)}
\onecolumn 
\maketitle

\appendix

\section{Additional Related Work}
\paragraph{Orthogonality constraints and disentanglement for generative models.}  In the context of generative adversarial networks (GANs) \citep{goodfellow2014generative}, the problem of analyzing and discovering interpretable directions has be studied recently by \citet{voynov2020unsupervised}. \citet{ren2022learning} propose a contrastive approach to discover interpretable directions using pretrained generative models. \citet{wei2021orthogonal} have proposed an orthogonality regularization of the Jacobian, which resulted in more interpretable generative abilities. \citet{ramesh2018spectral} constrain the right-singular vectors of a generator Jacobian to be unit directions, which corresponds to column-wise orthogonal generator Jacobians.  We go beyond these works by providing rigorous results on identifiability and by extending the scope to a encoder-only models.

\section{Proofs} \label{sec:app_proofs}

\subsection{Rotations Destroy Orthogonality Lemma}
We start by first proving an auxiliary lemma. We show that orthogonality of Jacobians, i.e., $\bm{J}_f\bm{J}_f^\top=\bm{S}$ with a diagonal matrix $\bm{S}$ will be destroyed in the general case when a rotation $\bm{R}$ is applied, such that $\bm{J}_{Rf}\bm{J}_{Rf}^\top = \bm{R}\bm{J}_f\bm{J}_f^\top\bm{R}^\top = \bm{R}\bm{S}\bm{R}^\top$ is not a diagonal matrix anymore.

\begin{lemma}[Rotations destroy orthogonality patterns.]
Let $\bm{S} \in \mathbb{R}^{K\times K}$ be a diagonal matrix, $\bm{S}=\diag{\left(\bm{s}\right)}$ with diagonal entries $\bm{s} > 0$ and $s_i \neq s_j, \forall i\neq j$, i.e., all diagonal entries of $\bm{S}$ are different and positive. Let $\bm{R} \in \mathbb{R}^{K\times K}$ be any rotation matrix with $\bm{R}^\top\bm{R}=\bm{I}$. If $\bm{R}\bm{S}\bm{R}^\top$ is a diagonal matrix, $\bm{R}$ must a signed permutation matrix (a permutation matrix where entries can be $\pm1$). \label{lem:helperorthogonality}
\end{lemma}

\textbf{Proof.} 
With $\bm{R}\bm{S}\bm{R}^\top = \diag\left(\lambda_1, \ldots \lambda_K \right)$, we have for each unit vector $\bm{e}^{(i)}$, $i = 1, \ldots, K$, that
\begin{align}
    \bm{R}\bm{S}\bm{R}^\top \bm{e}^{(i)} = \lambda_i \bm{e}^{(i)}\,\,.
\end{align}
We can represent $\bm{R}$ by its rows, $\bm{R}=\left[\bm{r}_1,\ldots, \bm{r}_K\right]^\top$ where each $\bm{r}_i\in \mathbb{R}^K$. In this notation, $\bm{R}^T \bm{e}^{(i)} = \bm{r}_i$, i.e., multiplication of the transpose with a unit vector will select the row $\bm{r}_i$. This results in \begin{align}
    \bm{R}\bm{S}\bm{r}_i = \lambda_i \bm{e}^{(i)}
\end{align}
Because $\bm{R}$ is invertible and square, we can left-multiply the equation by $\bm{R}^\top$. Using $\bm{R}^T \bm{e}^{(i)} = \bm{r}_i$ again, we arrive at
\begin{align}
    \bm{S}\bm{r}_i = \lambda_i \bm{r}_i.
\end{align}
This implies that all $\bm{r}_i$ are eigenvectors of the matrix $\bm{S}$ with the eigenvalues $\lambda_i$. By the initial assumption, $\bm{S}$ is a diagonal matrix with all-different entries $s_i$. The eigenvectors of such a matrix are only scaled unit vectors $\bm{e}^{(j)}$. Thus, each $\bm{r}_i$ will be a scaled unit-vector. The constraint of $\bm{R}$ being an orthogonal matrix enforces the $\bm{r}_i$ to be mutually different unit vectors with length $1$. Therefore, $\bm{R}$ necessarily has the form of a signed permutation. \hfill $\square$

Note that the converse is also true. If $\bm{R}$ is a signed permutation matrix, $\bm{R}\bm{S}\bm{R}^\top$ will be diagonal.

\subsection{PCA Ensures Identifiability (Theorem 3.1)}
\label{sec:proof_indep}

\begin{theorem}[PCA identifiability,~Theorem 3.1]
     Let $z_k, k=1, \dotsc, K,$ be uncorrelated random variables with non-zero and unequal variancecs. Let $\bm{e} = \bm{D}\bm{z}$, where $\bm{D} \in \mathbb{R}^{K \times K}$ is an orthonormal matrix. If an orthonormal post-hoc transformation $\bm{M}\in \mathbb{R}^{K \times K}$  results in mutually uncorrelated components $(z'_1, \dotsc, z'_K) = \bm{z}' = \bm{M}\bm{e}$, then $\bm{M}\ve = \bm{P}\bm{S} \vz$, where $\mP \in \mathbb{R}^{K \times K}$ is a permutation and $\bm{S} \in \mathbb{R}^{K \times K}$ is a matrix where $|s_{ii}|=1$ for $i \in 1,\ldots K$.
\end{theorem}

\textbf{Proof.} Since both $\mM$ and $\mD$ are orthogonal, $\bm{M}\bm{D}= \bm{Q}$ is also orthogonal. 
Our post-hoc transformation resulted in uncorrelated components, i.e., $\Covv(\bm{Q}\bm{x}) = \bm{Q}\Covv(\bm{x})\bm{Q}^\top\bm{\Gamma}$ is diagonal, where $\bm{\Gamma}$ is some diagonal matrix. Thus, $\bm{Q}\Covv(\bm{x})\bm{Q}^\top$ is diagonal, too.  We also know that our original components are uncorrelated with unequal variances, i.e., $\Covv(\bm{x}) = \diag(\bm{s})$ with $\bm{s} > 0$ and $s_i \neq s_j, \forall i\neq j$. Our helper \Cref{lem:helperorthogonality} then implies that $\bm{Q}$ must be a signed permutation. Thus, $\vz' := \bm{M}\ve = \mM \mD \vz = \mQ \vz =: \bm{P}\bm{S} \vz$, where $\mP \in \mathbb{R}^{K \times K}$ is a permutation and $\bm{S} \in \mathbb{R}^{K \times K}$ is a matrix where $|s_{ii}|=1$ for $i \in 1,\ldots K$. $\hfill \square$

\subsection{ICA Ensures Identifiability (Theorem 3.2)}

\begin{theorem}[ICA identifiability,~Theorem 3.2]
    Let $z_k, k=1, \dotsc, K,$ be independent random variables with non-zero variances where at most one component is Gaussian. Let $\bm{e} = \bm{D}\bm{z}$, where $\bm{D} \in \mathbb{R}^{K \times K}$ has full rank.  
    If a post-hoc transformation $\bm{M} \in \mathbb{R}^{N\times N}$ results in mutually independent components $(z'_1, \dotsc, z'_K) = \bm{z}' = \bm{M}\bm{e}$, then $\bm{M}\ve = \bm{P}\bm{S} \vz$, where $\mP \in \mathbb{R}^{K \times K}$ is a permutation and $\bm{S} \in \mathbb{R}^{K \times K}$ is a scaling matrix.
\end{theorem}

\textbf{Proof.} (1) We know that $\vz' = MDz =: \bm{C}' z$. Let us start with an additional assumption that both $\vz'$ and $\vz$ have unit variances. Then, by \citet[App. A .1]{comon1994independent}, $\bm{C}'$ must be orthonormal. 


Let us recall the following result
\begin{theorem}[Theorem 11 from \citet{comon1994independent}]
Let $\bm{x}$ be a vector with independent
components, of which at most one is Gaussian, and
whose densities are not reduced to a point-like mass.
Let $\bm{C}$ be an orthogonal $K \times K$ matrix and $\bm{z}$ the
vector $\bm{z} = \bm{C}\bm{x}$. Then the following three properties are equivalent:
\begin{enumerate}
    \item The components $z_i$ are pairwise independent.
    \item The components $z_i$ are mutually independent.
    \item $\bm{C} = \bm{S}\bm{P}$ where $\bm{S}$ is diagonal, $\bm{P}$ is a permutation.
\end{enumerate}
\end{theorem}

Since $\vz$ fulfills the conditions of this this theorem and $\bm{z}'$ has mutually independent entries, we know that $\bm{C}' = \bm{S}\bm{P}$.

(2) We now allow arbitrary variances, i.e., $\Covv(\vz')=\bm{\Lambda}$ and $\Covv(\vz)=\bm{\Gamma}$ where both covariance matrices are positive diagonal matrices. $\vz^\prime = \mM\mD \vz = \bm{C}^\prime \vz = \bm{\Lambda}^{1/2} \bm{\Lambda}^{-1/2} \bm{C}^\prime \bm{\Gamma}^{1/2} \bm{\Gamma}^{-1/2} \vz =: \bm{\Lambda}^{1/2} \bm{C}^{\prime\prime}  \bm{\Gamma}^{-1/2} \vz$. This is equivalent to $(\bm{\Lambda}^{-1/2} \bm{z}') = \bm{C}^{\prime\prime}  (\bm{\Gamma}^{-1/2} \vz)$. These rescaled random vectors both have unit variances, so (1) implies that $\bm{C}^{\prime\prime} = \bm{S'}\bm{P'}$. We can plug this back into the previous equation and see that $\vz^\prime = \bm{\Lambda}^{1/2} \bm{C}^{\prime\prime}  \bm{\Gamma}^{-1/2} \vz = \bm{\Lambda}^{1/2} \bm{S'}\bm{P'}  \bm{\Gamma}^{-1/2} \vz =: \bm{P'} \bm{S''} \vz$. Thus, $\vz' = \bm{M} \ve = \mM \mD \vz= \bm{P'}\bm{S''} \vz$, where $\mP' \in \mathbb{R}^{K \times K}$ is a permutation and $\bm{S''} \in \mathbb{R}^{K \times K}$ is a scaling matrix. $\hfill \square$

\subsection{Transfer lemma} \label{sec:prooftransfer}
DMA and IMA are based on structures in the Jacobian of the generative process. To be able to use them in the encoder and ultimately discover concepts, we first show that if an encoder mirrors the behavior of the generative process, up to a rotation and scale, its Jacobians must also mirror the Jacobians of the generative process.

\begin{lemma}[Transfer lemma] 
\label{lem:transferlemma}
Let $\vf$ be a faithful encoder for the generative process $\vg$ and further $\vf \circ \vg (\vz) = \mP \mS \vz$ $\forall \vz \in \mathcal{Z}$ where $\mP \in \mathbb{R}^{K \times K}$ is a permutation and $\mS \in \mathbb{R}^{K \times K}$ is a diagonal matrix. Then $\mJ_{\vf}(\vg(\vz)) = \mP' \mS' \mJ_\vg(\vz)^\top$  where $\mP' \in \mathbb{R}^{K \times K}$ is a permutation and $\mS' \in \mathbb{R}^{K \times K}$ is a diagonal matrix. 
\end{lemma}

\textbf{Proof.} Let $\vz \in \mathcal{Z}$ be arbitrary. $(\vf \circ \vg)(\vz) = \mP\mS\vz$ implies $\mJ_{\vf}(\vg(\vz)) \mJ_\vg(\vz) = \mP\mS$. Since $\vf$ is faithful to $\vg$,  $\mS$ has full rank, i.e., $\mS = \text{diag}(\alpha_1, \dotsc, \alpha_K)$ with $\alpha_k \in \mathbb{R}_{\neq 0}, k = 1, \dotsc, K$.

Now, let us write $\mJ_{\vf}(\vg(\vz)) = [\vvv_1, \dotsc, \vvv_K]^\top$ with $\vvv_i \in \mathbb{R}^{L}$. Similarly, we can write $\mJ_\vg(\vz) = [\vw_1, \dotsc, \vw_K]$ with $\vw_i \in \mathbb{R}^L, i = 1, \dotsc, K$.

Let us focus on an individual row of $\mJ_{\vf}$, i.e., let $k \in \{1, \dotsc, K\}$ be a fixed index of a row. Since $\mJ_{\vf}(\vg(\vz)) \mJ_\vg(\vz) = \mP \mS$ and $\mP$ is a permutation matrix with exactly one $1$ per row, there is precisely one column index $k'$ such that the $k$-th row and $k'$-th column of $\mP\mS$ is non-zero.
This setup allows drawing certain conclusions about the vector $\vvv_k$. Let $j = 1, \dotsc, K$ denote an arbitrary column of $\mP\mS$. Then,

(i) if $j = k'$, then $\vvv_k^\top \vw_{k'} = \alpha_{k'} \neq 0$. In consequence, $\vvv_k \neq 0$, $\vw_{k'} \neq 0$ and so we can decompose $\vvv_k = \va_k + \vb_k$, where $\va_k \in \text{span}(\{\vw_{k'}\}) \setminus \{0\}$ and $\vb_k \in \text{span}(\{\vw_{k'}\})^\bot$, where $^\bot$ denotes the orthogonal complement. Because $\text{span}(\{\vw_{k'}\}) = \left\{ \mu \vw_{k'}\middle| \mu \in \mathbb{R} \right\}$, we know that $\va_k=\frac{\alpha_{k'}}{\|\vw_{k'}\|^2_2}\vw_{k'}$.

(ii) if $j \neq k'$, then $\vvv_k^\top \vw_{j} = 0$. With (i), it follows that $\vb_k \in \text{span}\left(\{\vw_1, \dotsc, \vw_K\}\right)^\bot = \text{span}(\mJ_\vg(\vz))^\perp$. 

Since $\vf$ is faithful to $\vg$, we know that for each $\mathbf{c} \in \text{span}(\mJ_\vg(\vz))^\perp$, $\jacf(\vg(\vz))\mathbf{c} = \mathbf{0}$ ant therefore $\jacf(\vg(\vz))\bm{b}_k =\mathbf{0}$ This demands that the $k$-th component of the product is also 0, i.e., $\mathbf{v}_k\vb_k = (\va_k+\vb_k)^\top\vb_k=\va_k^\top\vb_k+ \vb_k^\top\vb_k=0$. By design $\va_k$ and $\vb_k$ are orthogonal such that immediately follows
$\bm{b}_k = \vzero$
Hence, $\vvv_k = \va_k + \vzero=\frac{\alpha_{k'}}{\|\vw_{k'}\|^2_2}\vw_{k'} + \vzero$ for our selected row $k$. Globally, this means $\mJ_{\vf}(\vg(\vz)) = \mP'\mS' \mJ_\vg(\vz)^\top$, with some scaling matrix $\mS'$ and permutation matrix $\mP'$. \hfill $\square$

\subsection{Disjoint Mechanisms ensure identifiability (Theorem 3.3)}\label{sec:app_proofdma}
\label{sec:proofident}
\begin{theorem}[Identifiability under DMA,~Theorem 3.3] 
Let $\vg$ have disjoint mechanisms and $\vf$ be a faithful encoder to $\vg$. 
If a full-rank post-hoc transformation $\bm{M} \in \mathbb{R}^{N\times N}$ results in disjoint rows in the Jacobian $\mM \mJ_{\vf}(\vg(\vz))$ for some $\vz \in \mathcal{Z}$, then $\bm{M}\ve = \bm{P}\bm{S} \vz$, where $\mP \in \mathbb{R}^{K \times K}$ is a permutation and $\bm{S} \in \mathbb{R}^{K \times K}$ is a scaling matrix.
\end{theorem}

\textbf{Proof.} We know that $\vf \circ \vg = \mD$ and $\mD$ has full rank. Since $\bm{M}$ also has full rank, there exists a non-singular matrix $\bm{E}'$ such that $\bm{M} = \bm{E}' \bm{D}^{-1}$. We can rewrite $\bm{E}' = \bm{S} \bm{E}$, where $\bm{E}$ has normalized rows and $\bm{S}$ is a diagonal matrix. 

Since $\bm{D}^{-1} \vf \circ \vg = \bm{I}$ and $\vg$ is DMA, we can apply the transfer lemma (\Cref{lem:transferlemma}). It implies that $\bm{D}^{-1} \jacf(g(z))$ has orthogonal rows. 

Suppose now for contradiction that $\bm{E}$ was not a permutation matrix. This means that without loss of generality the first row must contain at least two columns whose entries are not equal to zero. Since $\bm{E}$ has full rank, there must be a second row with a non-zero entry in at least one of these columns. Since $\bm{D}^{-1} \mJ_\vf(\vg(\vz_a))$ has disjoint rows, $\bm{S} \bm{E} \bm{D}^{-1} \mJ_\vf(\vg(\vz_a)) = \bm{M} \mJ_\vf(\vg(\vz_a))$ can no longer have disjoint rows. This contradicts the assumption. Hence, $\bm{E}$ must be a permutation matrix $\bm{P}$. This give $\bm{z}' = \mM \ve = \bm{P} \bm{S} \bm{D}^{-1} \bm{D} \vz = \bm{P} \bm{S} \vz$. \hfill $\square$

\subsection{Independent Mechanisms ensure Identifiability (Theorem 3.4)} \label{sec:app_proofima}
\begin{theorem}[Identifiability under IMA,~Theorem 3.4] 
Let $\vg$ adhere to IMA. Let $\vf$ be a faithful encoder to $\vg$. 
Suppose we have obtained an $\vf'= \mM \vf$ with a full-rank $\mM \in \mathbb{R}^{K \times K}$ and orthogonal rows in its Jacobian  $\jacfprime(\vg(\vz))$, i.e,  $\jacfprime(\bm{g}(\bm{z}))\jacfprime(\bm{g}(\bm{z}))^\top = \bm{\Sigma}(\bm{z})$ where $\bm{\Sigma}(\bm{z})$ is diagonal. If additionally for two points $\vz_a, \vz_b \in \mathcal{Z}$
and $\gamma_i \coloneqq \frac{\Sigma_{ii}(\vz_b)}{\Sigma_{ii}(\vz_a)}$ and $\forall i,j =1...K, i\neq j: \gamma_i \neq \gamma_j$ (NEMR condition), then $\bm{M}\ve = \bm{P}\bm{S} \vz$, where $\mP \in \mathbb{R}^{K \times K}$ is a permutation and $\bm{S} \in \mathbb{R}^{K \times K}$ is a scaling matrix.
\end{theorem}

\textbf{Proof.} We know that $\vf \circ \vg = \mD$ and $\mD$ has full rank. Since $\bm{M}$ also has full rank, there exists a non-singular matrix $\bm{E}$ such that $\bm{M} = \bm{E} \bm{D}^{-1}$. We will now show that the solution set of $\bm{E}$ can be constrained to be a permutation and scaling operation in three steps.

(1) $\mJ_{\vf^\prime}$ is orthogonal, i.e., $\bm{\Sigma}(\vz_a) = (\bm{M}\jacf(\vg(\vz_a))) (\bm{M}\jacf(\vg(\vz_a))^\top = ( \bm{E} \bm{D}^{-1} \jacf(g(z_a))) (\bm{E} \bm{D}^{-1} \jacf(g(z_a)))^\top = \bm{E} (\bm{D}^{-1} \jacf(g(z_a))) (\bm{D}^{-1} \jacf(g(z_a)))^\top \bm{E}^\top$. Since $\bm{D}^{-1} \vf \circ \vg = \bm{I}$ and $\vg$ is DMA, we can apply the transfer lemma (\Cref{lem:transferlemma}) and know that $\bm{D}^{-1} \jacf(g(z_a))$ must have orthogonal rows, i.e., $(\bm{D}^{-1} \jacf(g(z_a))) (\bm{D}^{-1} \jacf(g(z_a)))^\top = \bm{\Gamma}_a$, where $\bm{\Gamma}_a$ is some diagonal matrix with full rank. Substituting this back into the previous term, $\bm{\Sigma}(\vz_a) = \bm{E}  \bm{\Gamma}_a  \bm{E}^\top$.
The same holds for $\vz_b$, i.e., $\bm{\Sigma}(\vz_b) = \bm{E}  \bm{\Gamma}_b  \bm{E}^\top$.


(2) We've seen in (1) that both $\bm{\Sigma}(\vz_a)$ and $\bm{\Gamma}_a$ are the results of quadratic forms. Hence, their entries are all positive, and strictly positive because they have full rank. Thus we can define $\bm{Q} := \bm{\Sigma}(\vz_a)^{-1/2} \bm{E} \bm{\Gamma}_a^{1/2}$. Due to (1), $\bm{Q} \bm{Q}^\top = \bm{I}$, i.e., $\bm{Q}$ is orthogonal. It is easy to see that $\bm{E} = \bm{\Sigma}(\vz_a)^{-1/2} \bm{Q}\bm{\Gamma}_a^{1/2}$. In other words, $\bm{E}$ must be a (twice) scaled orthogonal matrix. 

(3) From (1) we get that
\begin{align}
    \bm{\Sigma}(\vz_a) \bm{\Sigma}(\vz_b)^{-1} &= \bm{E} \bm{\Gamma}_a \bm{E}^\top (\bm{E} \bm{\Gamma}_b \bm{E}^\top)^{-1} \\
    \bm{\Sigma}(\vz_a) \bm{\Sigma}(\vz_b)^{-1} &= \bm{E} \bm{\Gamma}_a \bm{\Gamma}_b^{-1} \bm{E}^{-1} \\
    \bm{E}^{-1} \bm{\Sigma}(\vz_a) \bm{\Sigma}(\vz_b)^{-1} \bm{E} &= \bm{\Gamma}_a \bm{\Gamma}_b^{-1} 
\end{align}
Now we can insert the result from (2)
\begin{align}
     \bm{\Gamma}_a^{-1/2} \bm{Q}^\top \bm{\Sigma}(\vz_a)^{1/2} \bm{\Sigma}(\vz_a) \bm{\Sigma}(\vz_b)^{-1} \bm{\Sigma}(\vz_a)^{-1/2} \bm{Q} \bm{\Gamma}_a^{1/2} &= \bm{\Gamma}_a \bm{\Gamma}_b^{-1} \\
     \bm{Q}^\top \bm{\Sigma}(\vz_a)^{1/2} \bm{\Sigma}(\vz_a) \bm{\Sigma}(\vz_b)^{-1} \bm{\Sigma}(\vz_a)^{-1/2} \bm{Q}  &= \bm{\Gamma}_a^{1/2} \bm{\Gamma}_a \bm{\Gamma}_b^{-1} \bm{\Gamma}_a^{-1/2} \\ 
    \bm{Q}^\top \bm{\Sigma}(\vz_a) \bm{\Sigma}(\vz_b)^{-1} \bm{Q}  &= \bm{\Gamma}_a \bm{\Gamma}_b^{-1} \\ 
\end{align}
Due to the NEMR condition, $\bm{\Sigma}(\vz_a) \bm{\Sigma}(\vz_b)^{-1}$ is a diagonal matrix with unequal positive entries. We can thus apply \Cref{lem:helperorthogonality} which implies that $\bm{Q} = \bm{P} \bm{S}$ where $\bm{P}$ is a permutation and $\bm{S}$ a diagonal matrix. Inserting this back into (2) gives $\bm{E} = \bm{\Sigma}(\vz_a)^{-1/2} \bm{Q}\bm{\Gamma}_a^{1/2} = \bm{\Sigma}(\vz_a)^{-1/2} \bm{P} \bm{S} \bm{\Gamma}_a^{1/2} = \bm{P} \bm{S}'$, where $\bm{S}'$ is a diagonal matrix. Hence, $\bm{z}' = \mM \ve = \bm{P} \bm{S}' \bm{D}^{-1} \bm{D} \vz = \bm{P} \bm{S}' \vz$.
\hfill$\square$

In the next section, we discuss how the proofs can be turned into analytical solutions to discover the ground truth components.
\subsection{Analytical Solutions to Concept Discovery}
\label{sec:app_analyticalsolutions}
\subsubsection{Disjoint Mechanisms}
Under a perfect DMA process $\vg$ and a noiseless faithful encoder $\vf$ to $\vg$, we can compute an analytical solution for $\mM$ that will result in an encoder $\vf'=\mM\vf$ that is compliant with the \textit{DMA criterion}, i.e., disjoint rows in its Jacobian. Suppose we are provided with a gradient matrix of $\vf$, $\jacf(\vx_a) \in \mathbb{R}^{K \times L}$. We propose the following steps:
\begin{enumerate}
    \item Select a submatrix  $\mJ_{reg} \in \mathbb{R}^{K \times K}$ of $K$ linearly independent columns in $\jacf(\bm{x}_a)$, such that  $\text{det}(\mJ_{reg}) \neq 0$.
    \item Compute and return $\mM = \mJ_{reg}^{-1}$
    \item This will result in $\vf^\prime= \mM\vf$ having disjoint rows in its Jacobian.
\end{enumerate}
\textbf{Proof}. $\jacf(\vx_a)$ must be of the form $\jacf(\vx_a) = \mH^{-1} \jacfstar(\bm{x}_a)$ for such an $\mM$ to exist, where $\jacfstar$ is the Jacobian of an encoder $\vf^{*}$ with disjoint rows and $\mH$ has full rank. $\mJ_{reg}$ can be written as $\mJ_{reg}=\mH^{-1} \mJ_{f^*, reg}$, where $\mJ_{f^*, reg}$ is a square submatrix of $\jacfstar$ with the same selected selected columns.
The submatrix $\mJ_{f^*, reg}$ also will be of to be of full rank because it can be written as $\mH\mJ_{reg}$, which are both full rank. Because of the DMA principle, $\mJ_{f^*, reg}$ again needs to be of the form $\mP \mS$ with one component active in each column. Furthermore, $\mM=\mJ_{reg}^{-1}=(\mH^{-1} \mP \mS)^{-1} = \mS^{-1}\mP^{-1}\mH$. As the inverses of scaling and permutation matrices have the same respective form again, $\mM\mH^{-1}=\mS^\prime\mP^\prime$. Therefore, $\vf^\prime = \mS^\prime\mP^\prime\vf^{*}$, maintaining its disjoint Jacobians.  

\subsubsection{Independent Mechanisms}
Suppose we are given matrices $\bm{\Sigma}(\vz_a)=\jacf(\bm{x}_a)\jacf(\bm{x}_a)^\top=\bm{D}^{-1}\bm{\Gamma}_a (\bm{D}^{-1})^\top$ and $\bm{\Sigma}(\vz_b)=\jacf(\bm{x}_b)\jacf(\bm{x}_b)^\top$.
We then apply the following steps
\begin{enumerate}
    \item $\bm{U} = \text{inverse}(\text{cholesky}(\bm{\Sigma}(\vz_a))$
    \item $\bm{V} = \text{eigenvectors}(\bm{U}\bm{\Sigma}(\vz_b)\bm{U}^\top)$
    \item return $\bm{H}=\bm{V}^\top\bm{U}$
\end{enumerate}
The first step implies that $\bm{U}^{-1}\bm{U}^{-\top}=\bm{\Sigma}(\vz_a)$ and that $\bm{U}\bm{\Sigma}(\vz_a) \bm{U}^\top=\bm{I}$. We have thus identified the matrix $\bm{E}$ from step (2) of the identifiability proof, which has the form $\bm{U}=\bm{\Lambda}^{1/2}\bm{Q}\bm{\Gamma}_a^{-1/2}\bm{M}$. In step two we compute $\bm{U}\bm{\Sigma}(\vz_b)\bm{U}^\top = \bm{\Lambda}^{1/2}\bm{Q}\bm{\Gamma}_a^{-1/2}\bm{\Gamma}_b\bm{\Gamma}_a^{-1/2}\bm{Q}^\top\bm{\Lambda}^{1/2} = \bm{V}\bm{R}\bm{V}^\top$, where $\bm{R}$ holds the eigenvalues. Accordingly, by left and right multiplying with $\bm{V}$, we observe that 
$(\bm{V}^\top\bm{U})\bm{\Sigma}(\vz_b)(\bm{V}^\top\bm{U})^\top = \bm{R}$, i.e., $(\bm{V}^\top\bm{U})$ solves the orthogonality problem for $\bm{\Sigma}(\vz_b)$. We can easily verify that $\bm{H}=\bm{V}^\top\bm{U}$ is also a solution for $\bm{\Sigma}(\vz_a)$ by computing $\bm{V}^\top\bm{U}\bm{\Sigma}(\vz_a) \bm{U}^\top\bm{V}=\bm{I}$. By the identifiability result, $\bm{H}=\bm{V}^\top\bm{U}=\bm{\Lambda}\bm{P}\bm{M}$, a scaled and permuted version of $\bm{D}^{-1}$, if the additional gradient ratio condition is fulfilled with $\vx_a$ and $\vx_b$.

\newcommand\mycommfont[1]{\footnotesize\ttfamily\textcolor{blue}{#1}}

\SetCommentSty{mycommfont}
\subsection{Algorithms}
\begin{algorithm}[t]
\DontPrintSemicolon
\caption{DMA concept discovery with SGD.\label{alg:da}}
\textbf{Input:} encoder $\vf$, images $\{\vx_n\}_{n=1, \dotsc, N}$\;
Jacobians $\gets$ Gradient($\vf$, $\{\vx_n\}_{n=1, \dotsc, N}$).detach()\;
$M \gets$ $K$-dim identity matrix\;
\For{$L$ \text{epochs}, $\mJ_\vf(\vx) \in$ \text{Jacobians}}{
    $\mU$ $\gets | \mM \mJ_\vf(\vx) |$\tcp*{No absolute value operation here for IMA}
    $\mU$ $\gets$ row-normalize $\mU$\;
    loss $\gets || \mU\,\mU^\top - \mI_K ||_F$\;
    loss.backwards() \tcp*{Optimize $M$}
}
\Return{$M$}
\end{algorithm}

\begin{algorithm}[t]
\DontPrintSemicolon
\caption{DMA concept discovery with SGD (determinant loss).\label{alg:dadet}}
\textbf{Input:} encoder $\vf$, images $\{\vx_n\}_{n=1, \dotsc, N}$\;
Jacobians $\gets$ Gradient($\vf$, $\{\vx_n\}_{n=1, \dotsc, N}$).detach()\;
$M \gets$ $K$-dim identity matrix\;
\For{$L$ \text{epochs}, $\mJ_\vf(\vx) \in$ \text{Jacobians}}{
    $\mU$ $\gets | \mM \mJ_\vf(\vx) |$\tcp*{No absolute value operation here for IMA}
    $\mV$ $\gets \mU\mU^\top$\;
    loss $\gets \log\left(\prod_i V_{ii}\right) - \log\text{det}\left(\mV\right)$\;
    loss.backwards() \tcp*{Optimize $M$}
}
\Return{$M$}
\end{algorithm}


We present the SGD optimization for DMA in \Cref{alg:da}. Note that the algorithm for IMA optimization via SGD can be obtained by just omitting the absolute value operation in the line indicated by the comment. \added{For the smaller toy datasets, we experiment with a version of the algorithm that uses the determinant (see \Cref{alg:dadet}), similar to the objective put forward by \cite{gresele2021independent}}.
As the determinant operation is hard to backpropagate through and might be unstable, we recommend \Cref{alg:da} for real-world applications and observed no significant performance differences on the datasets studied in this work.


\subsection{Extending gradients to general attributions}
\label{sec:app_gradtoattrib}
We make an initial attempt to generalize our method, considering gradients as a simple form of attribution method. Intuitively, $\mJ_\vf = \nabla_{\vx} (\vf(\vx))$ contains input gradients (termed grad in the remainder) which can be thought of as a simple form of attribution for each component \citep{simonyan2013deep, shah2021doinput}. 
Thus, on a more general level, our proposed approach optimizes for the disjointness of attributions. 
Thus, we may use other forms of \textit{homogeneous attributions} in place of $\mJ_\vf$. These are local attribution methods $A_\vf: \mathbb{R}^L \rightarrow \mathbb{R}^{K \times L}$ for the encoder $\vf$ with $A_{\mM\vf}(\vx) = \mM A_\vf(\vx)$ that map an instance $\vx$ to a matrix of attributions for each latent dimension. Besides the above input gradients, this class contains other popular methods such as integrated gradients (IG) \citep{Sundararajan2017} and smoothed gradients (SG) \citep{Smilkov2017} (because these methods are linear in $\vf$). Thus, we can formulate a generalized \textit{disjoint attributions objective}:
\begin{align}
    \min_{\mM} ~& \sum_{n=1}^{N} \left|\left| \left| \overline{\mM A_f(x)} \right|\,\left|\overline{\mM A_f(x)}\right|^\top - I_K\right|\right|^2_F.
    \label{eqn:orthogonaloptattr}
\end{align}
We indicate the row-normalization operation by the overbar, and denote by $\lvert\cdot\rvert$ the element-wise absolute values operation. Without the absolute value operation this results in the \emph{independent attributions objective}.

\section{Experimental Details}
We report the most important implementation details for our experiments in this section. Please confer the actual implementation available online\footnote{\url{https://github.com/tleemann/identifiable_concepts}} for full information. 
\label{sec:app_expdetails}
\subsection{Synthetic datasets}
\begin{figure}[tb]
    \begin{subfigure}[b]{0.35\linewidth}\centering
    \includegraphics[width=\textwidth]{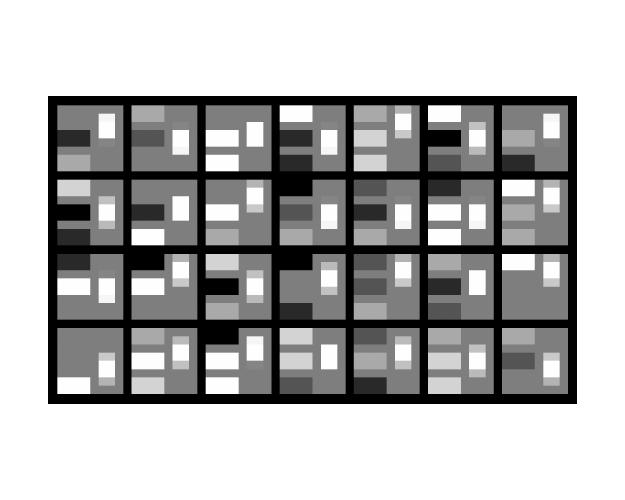}
    \caption{Random samples in the \texttt{FourBars} dataset.\newline~}
    \end{subfigure}
    \begin{subfigure}[b]{0.35\linewidth}
    \centering
    \includegraphics[width=\textwidth]{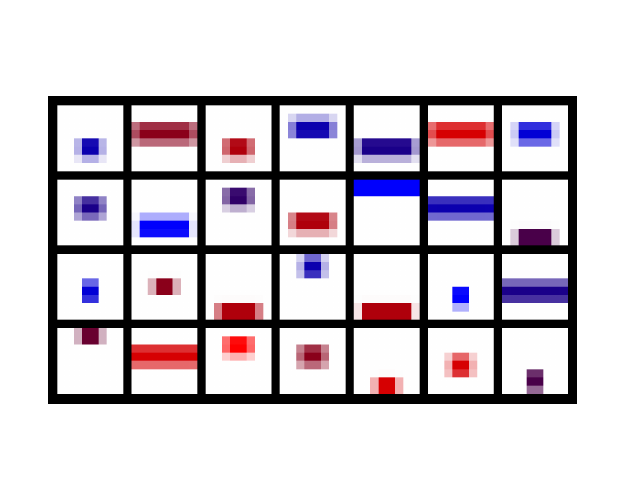}
    \caption{Random samples in the \texttt{ColorBar} dataset.\newline~}
    \end{subfigure}
    \begin{subfigure}[b]{0.24\linewidth}\centering
    \includegraphics[width=\textwidth]{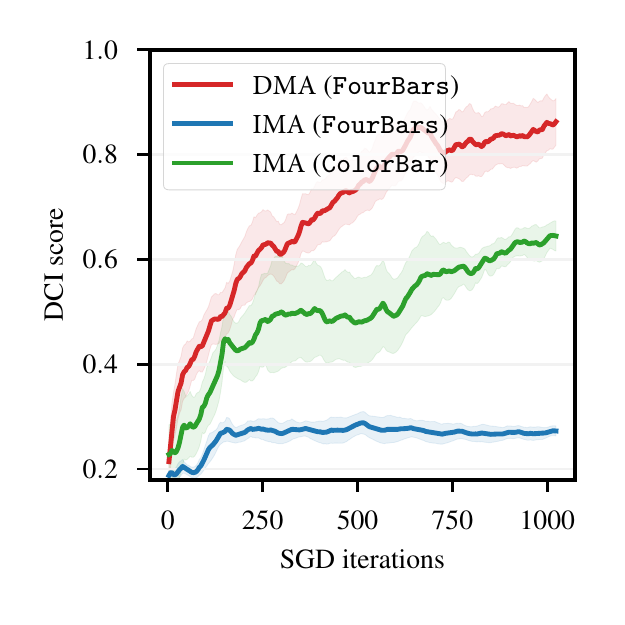}
    \caption{Disentangling gradients of synthetic datasets with SGD.\label{fig:sgdcurves}}
    \end{subfigure}
    \caption{Random samples drawn from the synthetic datasets (a,b). On the \texttt{FourBars} dataset, IMA fails to iterate towards a disentangled solution, because the non-equal magnitudes condition is violated. However, IMA converges on the \texttt{ColorBar} dataset, although at a slower rate (c)}
    \label{fig:randomsamplestoy}
\end{figure}

We show random samples from both datasets in \Cref{fig:randomsamplestoy}. We provide an additional graphics with the behavior on the synthetic datasets in \Cref{fig:sgdcurves}. They show that SGD exhibits a convergence behavior as predicted by our theory and comparable to the analytical solutions (shown in the main paper).

\subsection{Architectures} \label{sec:app_architectures} 
For the disentanglement models, we use the implementations provided by the open source library \texttt{disentanglement-pytorch}\footnote{\url{https://github.com/amir-abdi/disentanglement-pytorch}}. For the evaluation measures, we use the implementation of \texttt{disentanglement\_lib}\footnote{\url{https://github.com/google-research/disentanglement_lib}} with their respective default parameters. We use a simple encoder and decoder architecture, that consists of five and six feed-forward convolutional layers respectively and relies on the ReLU activation function.

\subsection{Correlated sampling} \label{sec:app_corrsampling}
\begin{figure}[tb]
    \centering
    \includegraphics{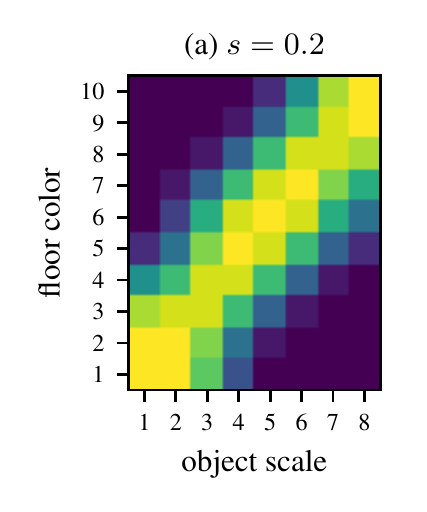}
    \includegraphics{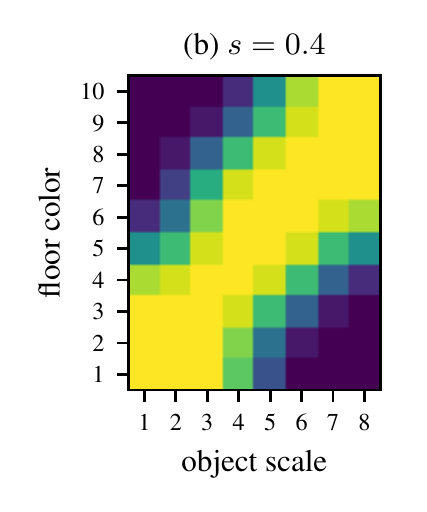}
    \includegraphics{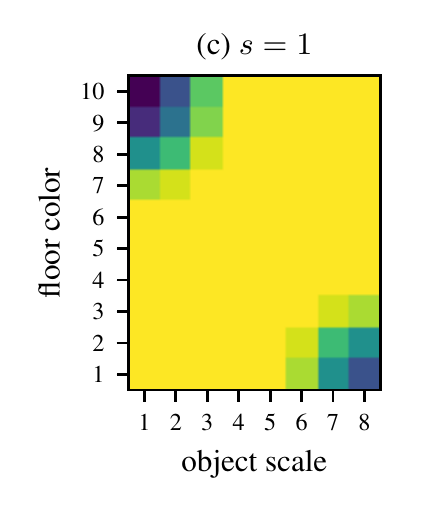}
    \includegraphics{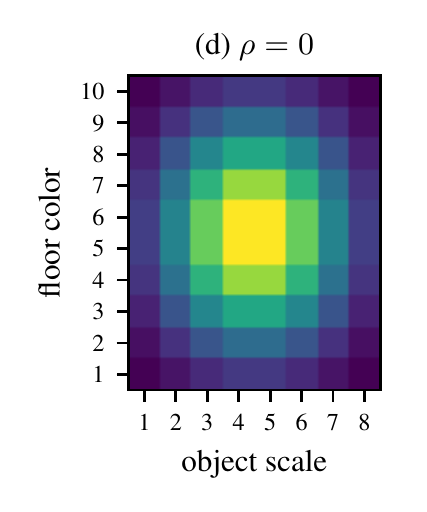}
    \includegraphics{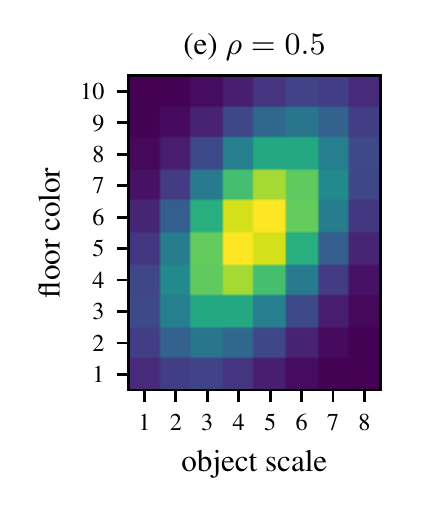}
    \includegraphics{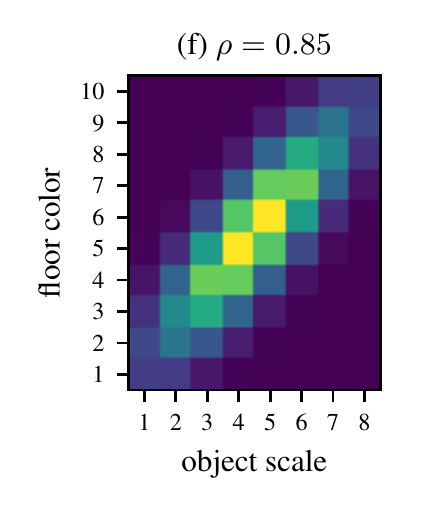}
    \caption{Exemplary correlated densities of the components floor color and object scale under the correlated sampling setup of \citet{gresele2021independent} (a -- c) and with our Gaussian sampling (d -- f). The correlation strength is indicated on top. Purple denotes a low and yellow a high density.}
    \label{fig:app_sample}
\end{figure}
In this paper, we use two methods to introduce correlations between the ground truth components. Both methods rely on proportional resampling: We first draw a batch that has multiple times the final batch size (we use factors from 3-6 depending on the non-uniformity of the distribution), then compute the (non-normalized) probability of each sample under a given distribution over the component values, and then resample a final batch (with replacement) proportional to these probabilities. 

The two methods differ in the probability distribution assigned to the component values. The first setting (used in \cref{sec:exp_comp}) uses the approach of \citet{trauble2021disentangled}: As visualized in \cref{fig:app_sample}(a) to (c), we pick two components $z_1$ and $z_2$, create the grid of possible values, and then lay a diagonal line over this grid. Along this line, we set a normal distribution with a standard deviation $s$. A higher $s$ means that the distribution gives a higher probability to more component combinations of the grid, whereas a smaller $s$ is more restrictive. Mathematically, it is defined by \citet{trauble2021disentangled} as:
\begin{align}
    p(z_1, z_2) \propto \exp\left(-\frac{(z_1 - \alpha z_2)^2}{2s^2}\right),
\end{align}
where $\alpha = z_1^{\text{max}} / z_2^{\text{max}}$ brings the components to a same scale and $s$ is similarly normalized to the maximum values that $z_1$ and $z_2$ can take. The remaining components $z_i, i > 2,$ are marginalized out of this distribution and thus continue to be sampled uniformly at random.

This setting is limited to one pair of components and also introduces a non-Gaussian distribution over all components. To tackle these limitations and thus to make the distributional challenge harder, we use a different probability distribution in \cref{sec:exp_robust}. Here, we lay a normal distribution over \emph{all} components, i.e., $z \sim \mathcal{N}(\mu, \Sigma)$, where $\mu$ is centered in the middle of the possible values, i.e., $\mu = \frac{z^{\text{max}} - z^{\text{min}}}{2}$. $\Sigma$ is similarly normalized, since we decompose it into $\Sigma = \text{diag}(\sigma^2) \Gamma$. The vector $\sigma \in \mathbb{R}_{>0}^K$ gives standard deviations for each component via $\sigma^2 = \left(\frac{\mu + 0.5}{2}\right)^2$ such that the distribution stretches across the grid of possible values. Note that the $+0.5$ is because the values are assumed to be zero-indexed. $\Gamma$ is a correlation matrix with $1$ on its diagonal. In the first experiment in \cref{sec:exp_robust}, we correlate only one pair of variables and set their corresponding off-diagonal entries in $\Gamma$ to $\rho$. \cref{fig:app_sample} (d) to (f) show the corresponding marginal distributions of these components. In the second experiment, we fill $\Gamma$ with several correlations in the following order:
\begin{align}\begin{matrix}
 z_1 \\
 z_2 \\
 z_3 \\
 z_4 \\
 z_5 \\
 z_6 \\
\end{matrix}
    \begin{pmatrix}
     & 1 & 4 & 12 & 14 & 9 \\
     & & 11 & 5 & 10 & 6 \\
     & & & 3 & 8 & 15 \\
     & & & & 13 & 7 \\
     & & & & & 2\\
     & & & & & & \\
    \end{pmatrix}
\end{align}
where the component order of the rows and columns is $z_1=$ \texttt{floor\_color}, $z_2=$ \texttt{background\_color}, $z_3=$ \texttt{object\_color}, $z_4=$ \texttt{object\_scale}, $z_5=$ \texttt{object\_shape}, $z_6=$ \texttt{orientation}. Here, it is important to ascertain that the covariance matrix stays positive definite. Thus, we start with $\rho = 0.7$, check if the lowest eigenvalue of $\Sigma$ is at least $0.2$, and if not, reduce $\rho$ by a factor of $0.9$ until the eigenvalue fulfills this property. While technically it would be enough to have the smallest eigenvalue anywhere above $0$, we found that $0.2$ helps in numerical stability, for instance when inverting the covariance matrix to compute the multivariate normal distribution density. 


\subsection{Discriminative setup} 
\label{sec:app_discrsetup} 
\begin{figure}[tb]
    \centering
    \includegraphics[width=\textwidth]{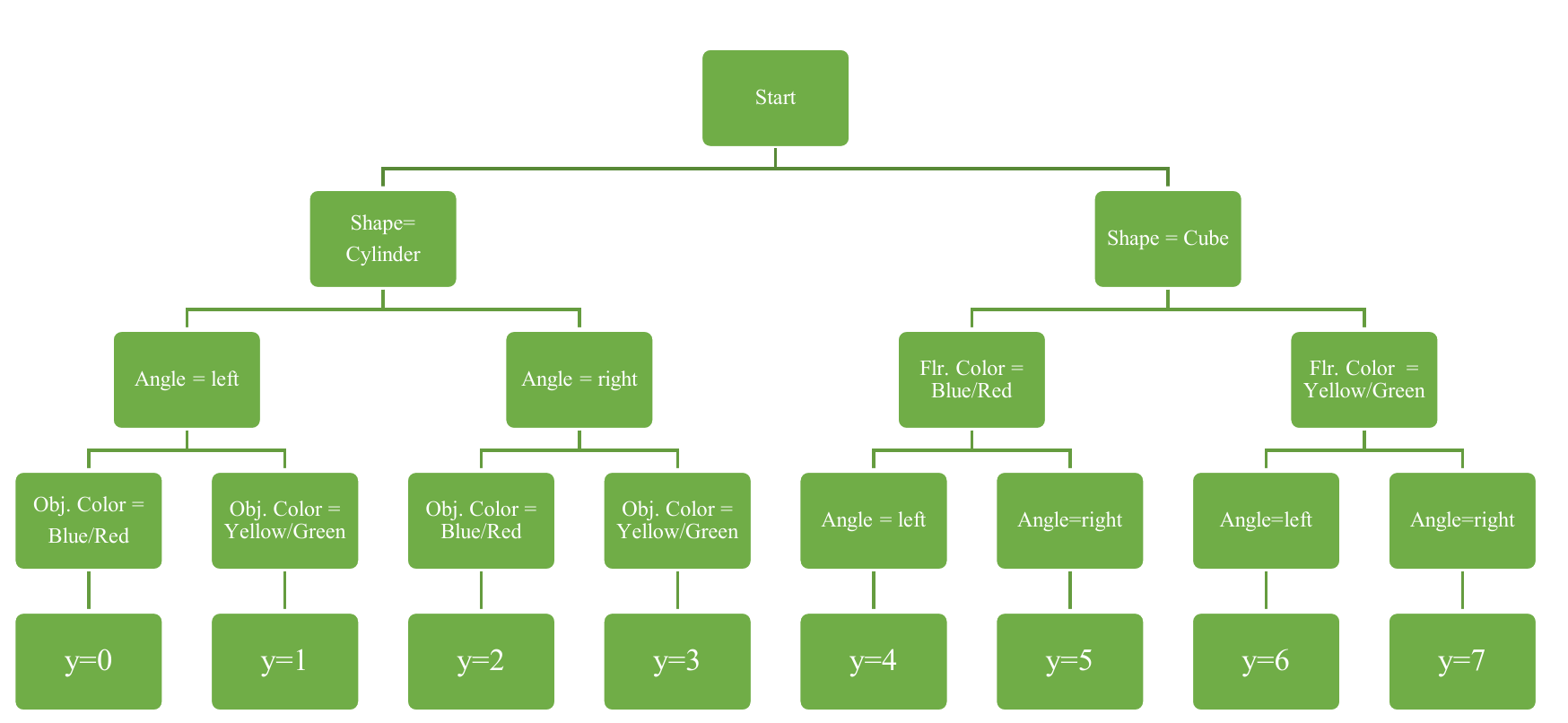}
    \caption{The decision tree setup that we use for the discriminative classification problem. Each image is assigned one out of eight class labels $y$ according to the following decision tree.}
    \label{fig:decisiontree}
\end{figure}

The decision tree that is used to generate the class distribution is shown in  \Cref{fig:decisiontree}. It relies on 4 (binarized) components. We trained a simple CNN classifier for this problem using the cross-entropy loss. In addition to the classification loss terms, we add a regularizer $\|\vz\|_2^2$, which constrains the latent codes to not grow arbitrarily large, during training. To create a realistic setup, we subsample the dataset to follow a normal distribution as shown in \cref{fig:app_sample}d. We also add label noise near the decision boundary: For objects which have an orientation that is nearly centered, we follow each branch (left/right) with a probability of 50\,\%. With increasing left-orientedness, the probability of following the left branch increases to almost 100\,\% in form of a sigmoid function over the actual orientation. We follow the same procedure for the remaining features. We train the classifier for 10k iterations at a batch size of 24 and verify that it reaches an accuracy close to the best-possible one taking the mislabeled samples into account. We add correlations by increasing the chance of the the factors \emph{obj. color} and \emph{floor color} taking the same binary value. We use our disjoint attributions approach to find a $H\in \mathbb{R}^{4\times 6}$ matrix that should map the 6-dimensional latent space of the model to the four binary concepts that are used in the classification task. For the unit directions, we take the first four unit directions of the latent space, for PCA and ICA, we take the most prominent four components discovered for the evaluation with the four annotated ground truth concepts.


\subsection{Evaluation scores} 
Several scores to quantify disentanglement have been proposed in the literature and often emphasize a different aspect of disentanglement \citep{sepliarskaia2019evaluating}. Among the most common scores is the Disentanglement-Completeness-Informativenss score (DCI) by \citet{eastwood2018framework}.  In their work, they propose a metric to measure Disentanglement, that relies on training predictors $\hat{z}_j = f_j(e)$  to predict each individual ground truth component $z_j$ from the learned latent representation $e$. Furthermore, they compute normalized importance weights $P_{ik}$  that quantify how important learned component $e_i$ is for predicting the ground component $z_k$. The disentanglement metric computes a row-wise entropy over the $P$-matrix, which assigns a score of 1, if the learned component $e_i$ is useful for predicting only a single factor and as score of 0, if it is equally useful for predicting all factors. Other commonly used metrics include the Mutual Information Gap (MIG) \citep{chen2018isolating}, Separated Attribute Predictability (SAP) \citep{kumar2018variational} and the FactorVAE metric \citep{kim2018disentangling}. However, it is unclear which of these metrics (or if any) also provide useful results in the correlated setting \citet{trauble2021disentangled}. Therefore, to compute the reliable evaluations, we train the model (and the post-processing methods such as PCA, ICA, IMA, DMA) on the correlated dataset, but compute the metrics on samples from the full, \emph{uncorrelated} datasets to avoid distortion in our scores. Träuble et al. noted that the DCI scores were able to discover entanglement between 2 variables \cite[Figure 11, Appendix]{trauble2021disentangled}, whereas most other metrics failed even in this case. Therefore, we mainly rely on this score for our experiments but also report results corresponding to \cref{sec:exp_comp} for the other scores that show a similar picture in this appendix (\cref{sec:app_furthermetricresults}).

\subsection{CUB experiments} \label{sec:app_cubdetails} 
CUB-200-2011 is a fine-grained dataset containing a total of 11,788 images of 200 bird species (5994 for training and 5794 for testing). We trained a ResNet-50 with two fully-connected (fc) layers (the second fc layer served as a bottleneck layer and took 2048-dim feature vectors as input and output 512-dim ones) on CUB for 100 epochs using a SGD optimizer with an initial learning rate of 0.001. The input images were center cropped to $224 \times 224$ pixels. Trained on a standard cross-entropy loss, the ResNet achieved a classification accuracy of on average 77.47\% on five random seeds, indicating proper training. After training the classifier, we applied our proposed method to discover components in the embedding space.

CUB provides no ground-truth components since it is a real-world dataset. It does, however, contain 312 attributes semantically describing the bird classes, e.g., wing color or beak shape. These attributes have no guarantee to be complete, but they offer $312$ interpretable components. 
This allows for an attempt to quantify whether our discovered components are interpretable and meaningful by comparing whether they match some of these interpretable ones.

Formally, we are given a set of image feature embeddings $\{\ve_n\}_{n = 1, \dotsc, N}$, $\ve_n \in \mathbb{R}^L$ and a  matrix $\mH = (\vh_1, \dotsc, \vh_K) \in \mathbb{R}^{L\times K}$ that contains the directions of discovered components ($L=512$, $K=30$). A score $s_n^k$ of $n$-th image for the $k$-th discovered component can be calculated by projecting the feature embeddings on that component direction, i.e., $s_n^k = \langle  \ve_n,  \vh_k  \rangle$. One pitfall is that $s_n^k$ can be negative, indicating, e.g., a non-black bird for the component "primary color: black", but this opposite attribute is usually encoded in a separate attribute in CUB, e.g., "primary color: white". Thus, we separate the negative and positive values into two components (where we set values of the opposite sign to 0), resulting in $2\cdot K$ positive scores for each image. 

To compare these component scores with the attributes, we make use of the numerical attribute values provided in CUB. First, we average the $2 \cdot K$ component values of all images of a class, to be comparable with the class-wise attributes provided by CUB. This gives us a numerical $2 \cdot K$ dimensional component description and a $312$ dimensional attribute description per class. Now, we match the discovered components to the attributes. We compare each discovered component to each attributes via the Spearman's rank correlation coefficient and consider the attribute with the highest score to match the component. These are the matches used in \cref{sec:exp_cub}. We further use the (average) Spearman's rank correlation across all components to their best-matching attributes to quantify how well the components match to interpretable attributes in \cref{sec:app_cub_eval}.

\subsection{Hyperparameters for the disentanglement models} \label{sec:app_hyperparameters} 
We orient our hyperparameter ranges by the works of \citet{trauble2021disentangled, locatello2019challenging}. The exact ranges are provided in \cref{tab:hyperparameterranges}. We find the best hyperparameters in the ranges for each correlation strength/dataset/model triple separately. Then we train five models from independent seeds to run our experiments.
We use the Adam optimizer for all model with a learning rate of $10^{-4}$, batch size of 64 and train for 300k iterations (equiv. to 40 epochs on Shapes3D). 

For the optimization of the post-hoc disentanglement problem, we use slightly different hyperparameters. We use the RMSProp optimizer with learning rate of $10^{-3}$ and a batch size of 48.

\begin{table}[tb]
    \centering
    \begin{tabular}{cc}
    \toprule
        Model & Ranges \\
    \midrule
        BetaVAE & $\beta \in \{1,2,4,6,8,16\}$ \\
        FactorVAE & $\gamma \in \{5, 8, 10, 20, 30, 40, 50, 100\}$\\
        BetaTCVAE & $\beta \in \{1,2,4,6,8,10\}$ \\
        DIPVAEI & $\lambda_{od} \in \{1,2,5,10, 20, 50\}$ \\
    \bottomrule
    \end{tabular}
    \caption{The hyperparameter ranges considered in this work.}
    \label{tab:hyperparameterranges}
\end{table}

\begin{table}[tb]
\centering
\adjustbox{width=0.5\textwidth}{
\begin{tabular}{r*{3}{c}}
\toprule
 Dataset & \multicolumn{3}{c}{MPI3D-real}\\
\cmidrule{1-1} \cmidrule(lr){2-4}
 \wrapb{Correlated}{components} &  \wrapb{background \&}{object color} &  \wrapb{background \&}{robot arm dof-1} &  \wrapb{robot arm dof-1 \&}{robot arm dof-2}\\
\cmidrule{1-1} \cmidrule(lr){2-4} 
\textbf{BetaVAE} &
\res{0.340}{0.027} & \res{0.277}{0.026} & \res{0.300}{0.046} \\
+PCA &
\res{0.116}{0.008} & \res{0.174}{0.021} & \res{0.154}{0.015}\\
+ICA &
\res{0.237}{0.042} & \res{0.205}{0.023} & \res{0.180}{0.021} \\
+Ours (IMA) & 
\bres{0.355}{0.033} & \bres{0.349}{0.015} & \bres{0.337}{0.038} \\
+Ours (DMA) &
\res{0.334}{0.025} & \res{0.317}{0.028} & \res{0.278}{0.030} \\
\cmidrule{1-1} \cmidrule(lr){2-4} 

\textbf{FactorVAE} &
\bres{0.205}{0.022} & \bres{0.239}{0.017} & \res{0.171}{0.005}\\
+PCA&
\res{0.179}{0.010} & \res{0.234}{0.012} & \res{0.171}{0.006}\\
+ICA&
\res{0.066}{0.009} & \res{0.090}{0.006} & \res{0.073}{0.011}\\
+Ours (IMA)&
\res{0.201}{0.019} & \res{0.226}{0.010} & \bres{0.191}{0.011}\\
+Ours (DMA)&
\res{0.184}{0.013} & \res{0.218}{0.016} & \res{0.180}{0.013}\\
\cmidrule{1-1} \cmidrule(lr){2-4} 

\textbf{BetaTCVAE} &
\bres{0.383}{0.022} & \bres{0.359}{0.026} & \bres{0.309}{0.036}\\
+PCA&
\res{0.356}{0.022} & \res{0.328}{0.017} & \res{0.295}{0.038}\\
+ICA& 
\res{0.245}{0.041} & \res{0.260}{0.024} & \res{0.170}{0.045}\\
+Ours (IMA)&
\res{0.323}{0.025} & \res{0.316}{0.029} & \res{0.271}{0.033} \\
+Ours (DMA)&
\res{0.327}{0.027} & \res{0.325}{0.025} & \res{0.272}{0.033} \\
\cmidrule{1-1} \cmidrule(lr){2-4}

\textbf{DipVAE} &
\res{0.235}{0.019} & \res{0.181}{0.049} & \res{0.232}{0.040}\\
+PCA&
\res{0.090}{0.005} & \res{0.088}{0.028} & \res{0.091}{0.011}\\
+ICA&
\res{0.234}{0.019} & \res{0.180}{0.048} & \res{0.232}{0.041}\\
+Ours (IMA)&
\res{0.230}{0.022} & \res{0.182}{0.048} & \res{0.230}{0.042} \\
Ours (DMA)&
\bres{0.249}{0.026} & \bres{0.188}{0.049} & \bres{0.253}{0.051}\\
 \bottomrule
  \end{tabular}
}
\caption{MPI-3D dataset: Mean $\pm$ std. err. of the DCI scores (across all components of the dataset) of several models and post-hoc methods applied to their embeddings. Columns show which pair of components was correlated during training.\label{tab:app_posthocdisentangle}}
\end{table}
\subsection{Details on the introductory example}\label{sec:app_detailsintrofig}
\begin{figure*}[t]
\centering
\includegraphics[width=0.9\textwidth]{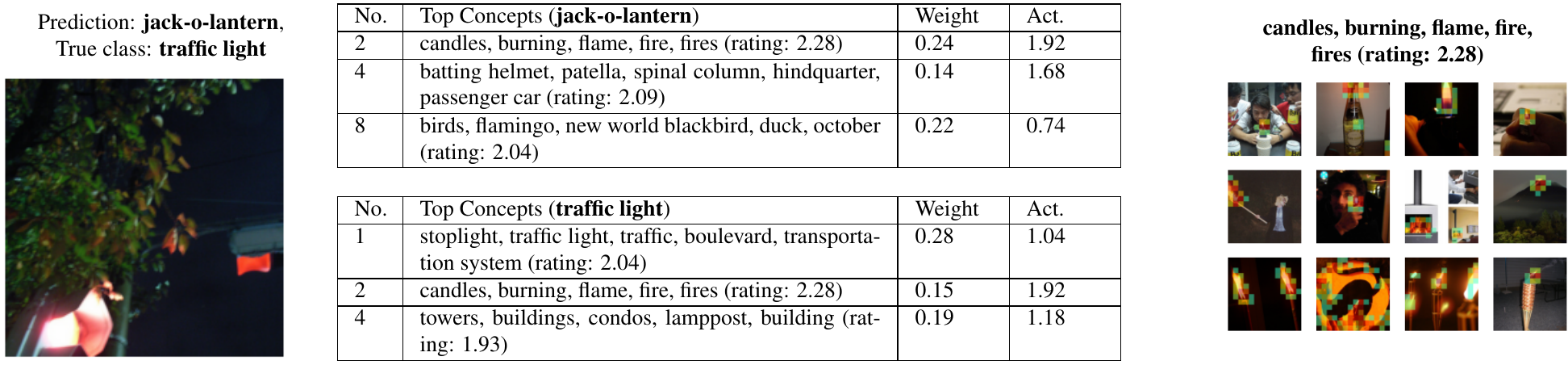}
\caption{Original local conceptual explanation of the missclassification. We find that the most activating concept ``candles, burning, flame...'' activates for very dark images. This concept is also highly activated for the traffic light example. We cleared up the description of the concepts for the motivational figure.}
\label{fig:localexplwecann}
\end{figure*}

The introductory example is inspired by a real explanation generated for a missclassification of the ResNet50 model pretrained on the ImageNet \citep{russakovsky2015imagenet} dataset delivered with the popular \texttt{pytorch} \citep{paszke2017automatic} package. Using the approach devised by \cite{leemann2022coherence}, we use the individual neurons of the classifier's last-layer as concepts and describe them by words. We obtain the conceptual explanation shown in \Cref{fig:localexplwecann}. We simplify the explanation for the motivational figure and give the concepts relatable names. However, the gist of the example stays the same.

\section{Additional results}\label{sec:app_additionalresults}
\subsection{Reconstruction quality} \label{sec:app_reconstruction} 

\begin{figure}[tb]
    \centering
    \includegraphics[scale=0.35]{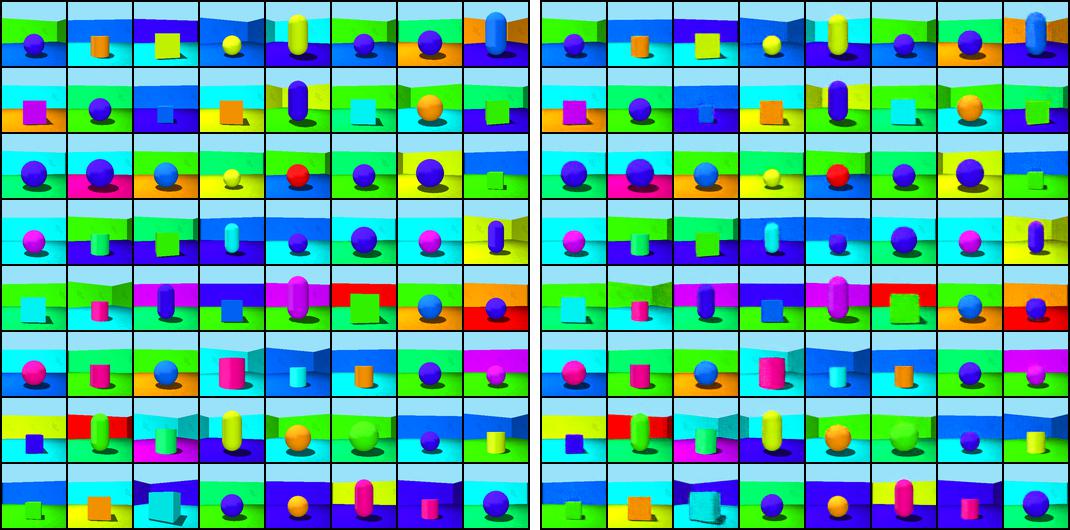}
    \caption{Random example images (left) and their reconstructions (right) of a BetaVAE on Shapes3D.}
    \label{fig:app_reconcub}
\end{figure}

\begin{figure}[tb]
    \centering
    \includegraphics[scale=0.35]{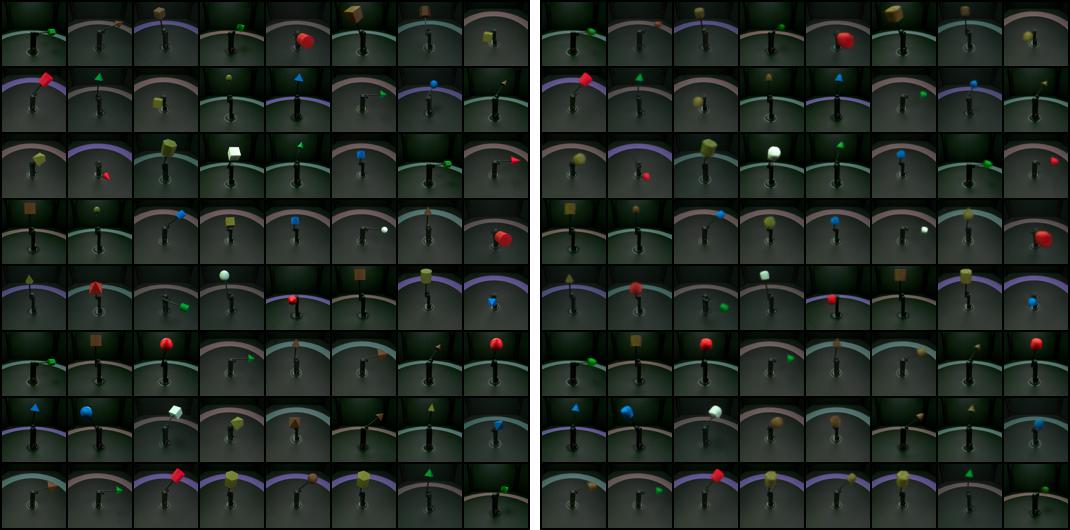}
    \caption{Random example images (left) and their reconstructions (right) of a BetaVAE on MPI3D-real.}
    \label{fig:app_reconmpi}
\end{figure}

As a check, we investigate the reconstruction quality of the disentanglement models. For the 3D shapes, the reconstruction is very high, but we observe some more serious reconstruction errors on the MPI-3d dataset (see \cref{sec:app_mpi3d}). %
Figures~\ref{fig:app_reconcub} and \ref{fig:app_reconmpi} show the original images on the left and the reconstructions of a randomly chosen BetaVAE on the right. On Shapes3D, the BetaVAE is able to reconstruct the image from its embedding representation. On MPI3D-real, it is able to reconstruct the big image parts shared across many pictures (ground, background stripe and background), but becomes blurry in the smaller and more nuanced robot arm and object shapes. This indicates that the information on these components might not be stored in the embedding space and is thus hardly disentanglable. A longer training (800k instead of 300k iterations) did not resolve the issue. The issue might arise, following \citet{gondal2019transfer}, because the input images were scaled down to 64x64 pixels making the detailed objects hard to perceive, and because the same architecture as in the Shapes3D experiments was used, which might not be expressive enough. 

\subsection{Results for the MPI-3D dataset}
\label{sec:app_mpi3d}
In addition to 3Dshapes, we use the challenging MPI3D-real dataset \citep{gondal2019transfer}, which consists of realistic images of a moving robot arm. It is by far more challenging, as the component is only present in a small portion of the images, and the data consists of real photographs. We report the results on this dataset in \Cref{tab:app_posthocdisentangle}. We saw low disentanglement scores of both the base and post-hoc models on MPI3D-real compared to the performance on Shapes3D. This implies that the embedding spaces of the VAEs was not trained well. In fact, this is supported by the reconstruction quality considerations on both Shapes3D and MPI3D-real. Because our approaches are based on the given embeddings, they also struggle when they incorrectly reflect the sample. 

\subsection{Correlation strengths and attribution methods in first experiment} 
\label{sec:app_morerectify}
In this section we provide additional ablations for the rectification experiment in \cref{sec:exp_comp}. We investigate the impact of the choice of attribution method (\cref{sec:app_gradtoattrib}) and the correlation strength $s$. The values (DCI scores) are shown in \cref{tab:app_further_cor}. As expected, our approach offers the highest gains over the baseline when the correlation is higher. Starting at $s=0.4$, our runs start to reliably outperform the baselines. Regarding the attributions, there is no clear picture, but Grad and SG seem to yield good results more stably across runs. DMA usually outperforms IMA, which supports our theoretical results on identifiability.

\subsection{Further disentanglement metrics} 
\label{sec:app_furthermetricresults}
Tables \ref{tab:app_posthocdisentangle_mig} -- \ref{tab:app_posthocdisentangle_sap} show the results of the experiment in \cref{sec:exp_comp} measured in the alternative metrics MIG, FactorVAE and SAP score. 
For MIG, we see similar results as for DCI in \Cref{tab:posthocdisentangle} and in \Cref{tab:app_posthocdisentangle}. The results in FactorVAE and SAP score are slightly inferior but our approach still improves over the baseline in many setups. We also compute the disentanglement only on the two correlated components for the first pair of factors in \Cref{tab:app_pairwisedci}. This emphasized the improvement introduced by our IMA and DMA approaches.

\subsection{Runtimes and further ablation studies}\label{sec:app_ablations}
\added{\textbf{Runtime.} Runtime can be an important concern for algorithms in explainable AI, for instance when they are to be deployed on embedded devices. We therefore report the runtimes required to obtain the results shown in \Cref{tab:posthocdisentangle} here:}
\begin{center}
\begin{tabular}{ccccc}
\toprule
Algorithm & PCA	& ICA & Ours-DMA & Ours-IMA \\
\midrule
Runtime (sec) &  316 $\pm$ 38 & 320 $\pm$ 44 & 1140 $\pm$ 97	& 1017 $\pm$ 121 \\
\bottomrule
\end{tabular}
\end{center}

\added{For our SGD-based optimization, we note that the user can choose how many optimization steps are executed. In the present work, we chose 20000 steps to make sure that the optimization has converged. Using these settings, the runtime of our algorithms is approximately 3 times as high as that of the baseline. We think that this is not prohibitively more expensive. However, convergence of the optimization is usually achieved much quicker.}

\added{\textbf{Effect of less SGD iterations.} To ablate the behavior of our approach with a smaller runtime budget, we rerun all the approaches in \Cref{tab:posthocdisentangle} using only 8000 iterations, making the runtime approximately equal across methods. We report the DCI scores as in the original table in \Cref{tab:app_lessiterations} and see that our DMA approach still outperforms all the baselines in 10 of 12 settings. Thus, even when runtime is an important concern in the evaluation, our approach can still yield competitive results.}

\added{\textbf{Robustness with respect to noise.} While IMA covers a more general class of functions, we empirically observed superior performance for DMA in most experiments. We therefore hypothesize that the performance difference stems from the behavior of IMA and DMA under noisy gradients and from the approximate optimizers that we use. We conduct an ablation study to obtain further evidence for these hypotheses. We modify the \texttt{FourBars} dataset to fulfill NEMR by adding varying magnitudes of the component gradients in the rows of $\jacf(\vg(\vx))$. This dataset is now solvable by both IMA and DMA. We then add noise to the analytical gradients. We perform a fixed number of 500 SGD steps of \Cref{alg:dadet} and otherwise use the same optimizer parameters as in the main paper. We obtain the DCI curves across different noise levels shown in \Cref{fig:app_robustness}. Without noise, both algorithms find disentangled solutions with DCI scores >0.9 (practically perfect disentanglement when evaluated on traversals). When we add noise, the disentanglement scores decrease as the working assumptions now only hold approximately. At a noise level of 0.1, the actual gradients shown in \Cref{fig:fourbars} are hard to see already with bare eyes. At each point there is a small but consistent gap between the performance of IMA and DMA, indicating that the DMA objective often finds better solutions with the standard SGD optimizer pipeline. This matches our empirical findings of the real data experiments.
}


\begin{figure}[tb]
\begin{minipage}[b]{\textwidth}
\begin{minipage}[t]{0.6\textwidth}
\scalebox{0.75}{
\begin{tabular}{r*{6}{c}}
\toprule
\wrapb{Correlated}{components} & \multicolumn{2}{c}{\wrapb{floor \&}{background}}  &\multicolumn{2}{c}{\wrapb{orientation \&}{background}} &  \multicolumn{2}{c}{\wrapb{orientation \&}{size}}\\
\cmidrule{1-1}\cmidrule(lr){2-3}\cmidrule(lr){4-5} \cmidrule(lr){6-7}
\textbf{BetaVAE} &\res{0.497}{0.03}& & \res{0.581}{0.04}& & \res{0.491}{0.05}& \\
+PCA&\res{0.263}{0.03}&\negimp{-47\%}& \res{0.310}{0.02}&\negimp{-47\%}& \res{0.324}{0.04}&\negimp{-34\%}\\
+ICA&\bres{0.574}{0.04}&\posimp{+16\%}& \res{0.540}{0.08}&\negimp{-7\%}& \res{0.577}{0.04}&\posimp{+17\%}\\
+Ours (OA)&\res{0.533}{0.11}&\posimp{+7\%}& \res{0.594}{0.04}&\posimp{+2\%}& \res{0.576}{0.03}&\posimp{+17\%}\\
+Ours (DA)&\res{0.472}{0.14}&\negimp{-5\%}& \bres{0.633}{0.05}&\posimp{+9\%}& \bres{0.617}{0.03}&\posimp{+26\%}\\
\cmidrule{1-1} \cmidrule(lr){2-6}
\textbf{FactorVAE} &\res{0.507}{0.11}& & \res{0.502}{0.08}& & \bres{0.712}{0.01}& \\
+PCA&\res{0.358}{0.07}&\negimp{-29\%}& \res{0.474}{0.05}&\negimp{-5\%}& \res{0.556}{0.03}&\negimp{-22\%}\\
+ICA&\res{0.294}{0.07}&\negimp{-42\%}& \res{0.263}{0.05}&\negimp{-48\%}& \res{0.340}{0.03}&\negimp{-52\%}\\
+Ours (OA)&\res{0.539}{0.04}&\posimp{+6\%}& \res{0.498}{0.03}&\negimp{-1\%}& \res{0.568}{0.06}&\negimp{-20\%}\\
+Ours (DA)&\bres{0.567}{0.07}&\posimp{+12\%}& \bres{0.531}{0.04}&\posimp{+6\%}& \res{0.571}{0.02}&\negimp{-20\%}\\
\cmidrule{1-1} \cmidrule(lr){2-6}
\textbf{BetaTCVAE} &\res{0.619}{0.01}& & \res{0.613}{0.04}& & \res{0.659}{0.01}& \\
+PCA&\res{0.400}{0.03}&\negimp{-35\%}& \res{0.421}{0.07}&\negimp{-31\%}& \res{0.450}{0.07}&\negimp{-32\%}\\
+ICA&\res{0.540}{0.02}&\negimp{-13\%}& \res{0.497}{0.04}&\negimp{-19\%}& \res{0.627}{0.02}&\negimp{-5\%}\\
+Ours (OA)&\res{0.635}{0.04}&\posimp{+3\%}& \res{0.648}{0.03}&\posimp{+6\%}& \res{0.682}{0.02}&\posimp{+4\%}\\
+Ours (DA)&\bres{0.644}{0.01}&\posimp{+4\%}& \bres{0.659}{0.02}&\posimp{+8\%}& \bres{0.724}{0.02}&\posimp{+10\%}\\
\cmidrule{1-1} \cmidrule(lr){2-6}
\textbf{DipVAE} &\res{0.631}{0.02}& & \res{0.652}{0.02}& & \res{0.548}{0.04}& \\
+PCA&\res{0.158}{0.01}&\negimp{-75\%}& \res{0.160}{0.02}&\negimp{-75\%}& \res{0.170}{0.02}&\negimp{-69\%}\\
+ICA&\res{0.630}{0.02}&\negimp{-0\%}& \res{0.651}{0.02}&\negimp{-0\%}& \res{0.542}{0.03}&\negimp{-1\%}\\
+Ours (OA)&\res{0.640}{0.01}&\posimp{+1\%}& \res{0.621}{0.02}&\negimp{-5\%}& \res{0.545}{0.05}&\negimp{-1\%}\\
+Ours (DA)&\bres{0.683}{0.01}&\posimp{+8\%}& \bres{0.676}{0.01}&\posimp{+4\%}& \bres{0.591}{0.06}&\posimp{+8\%}\\
 \bottomrule
 \end{tabular}}

\captionof{table}{Using 8000 instead of 20000 SGD iterations: Mean $\pm$ std. err. of the DCI scores of post-hoc methods applied to the embedding spaces of four disentanglement architectures with different pairs of correlated variables. Our DMA method still yields competitive results even with fewer SGD steps.\label{tab:app_lessiterations}}
\end{minipage}
\hfill
\begin{minipage}[t]{0.37\textwidth}
\centering
\scalebox{0.8}{
\begin{tabular}{r*{1}{c}}
\toprule
 Dataset & \multicolumn{1}{c}{Shapes3D} \\
\cmidrule{1-1} \cmidrule(lr){2-2}
 \wrapb{Correlated}{factors} & \wrapb{floor vs.}{background} \\
\cmidrule{1-1} \cmidrule(lr){2-2} 
\textbf{BetaVAE} &\res{0.579}{0.089} \\
+PCA&\res{0.291}{0.033} \\
+ICA&\res{0.435}{0.076} \\
+IMA-SGD&\ures{0.738}{0.072} \\
+DMA-SGD&\bres{0.868}{0.025} \\
\cmidrule{1-1} \cmidrule(lr){2-2}
\textbf{FactorVAE} &\res{0.684}{0.163} \\
+PCA&\res{0.526}{0.136} \\
+ICA&\res{0.363}{0.097} \\
+IMA-SGD&\ures{0.779}{0.063} \\
+DMA-SGD&\bres{0.847}{0.072} \\
\cmidrule{1-1} \cmidrule(lr){2-2}
\textbf{BetaTCVAE} &\res{0.589}{0.005} \\
+PCA&\res{0.388}{0.046} \\
+ICA&\res{0.609}{0.065} \\
+IMA-SGD&\bres{0.876}{0.027} \\
+DMA-SGD&\ures{0.754}{0.127} \\
\cmidrule{1-1} \cmidrule(lr){2-2}
\textbf{DipVAE} &\res{0.615}{0.114} \\
+PCA&\res{0.429}{0.169} \\
+ICA&\res{0.585}{0.024} \\
+IMA-SGD&\bres{0.798}{0.099} \\
+DMA-SGD&\ures{0.782}{0.009} \\
 \bottomrule
  \end{tabular}}
\captionof{table}{Mean $\pm$ std. err. of the DCI scores of four post-hoc methods applied to the embedding spaces of four disentanglement models on two datasets with different pairs of correlated variables. The DCI is computed across \textbf{the two correlated components} of the dataset.\label{tab:app_pairwisedci}}
\end{minipage}
\end{minipage}
\end{figure}
\begin{figure}[tb]
    \centering
    \includegraphics[width=0.5\textwidth]{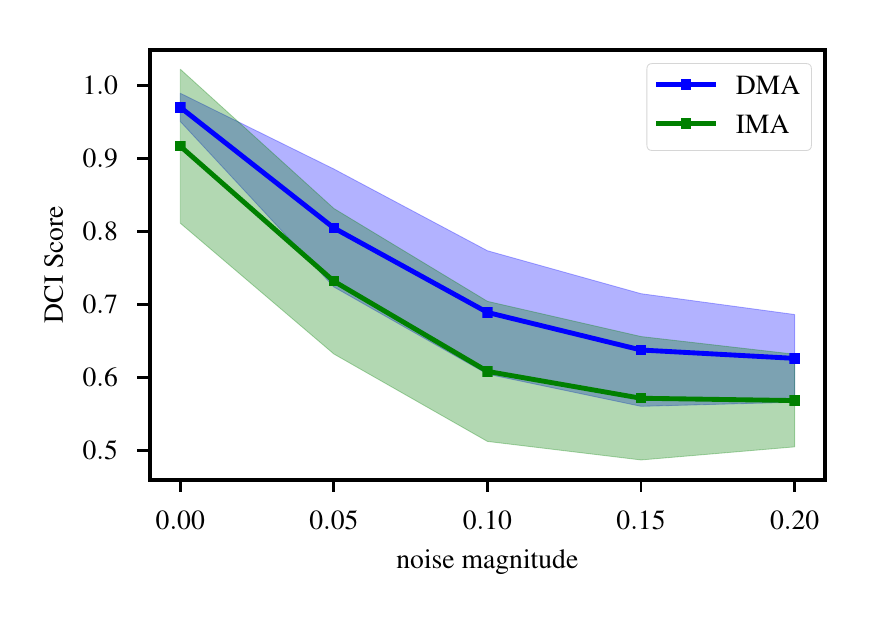}
    \caption{Robustness of optimization to noisy gradients. We use a variant of the \texttt{FourBars} dataset that can be identified both by IMA and DMA (the NEMR condition holds) and add noise of increasing magnitude to the analytical gradients. While the disentanglement scores (DCI) decrease for both methods, we observe that the performance of IMA under noise is slightly worse than that of DMA. This may be one factor contributing to the weaker overall performance of IMA as compared to DMA.\label{fig:app_robustness}}
   
\end{figure}
\newcommand{\wres}[2]{\begin{tabular}[c]{@{}c@{}}#1 \\ \small{$\pm$#2} \end{tabular}}
\newcommand{\bwres}[2]{\begin{tabular}[c]{@{}c@{}}\textbf{#1} \\ \small{$\pm$\textbf{#2}} \end{tabular}}

\begin{table}[tb]
\adjustbox{width=\textwidth}{
\begin{tabular}{r*{12}{c}}
\toprule
 Model & \multicolumn{3}{c}{BetaVAE} & \multicolumn{3}{c}{FactorVAE} & \multicolumn{3}{c}{BetaTCVAE} & \multicolumn{3}{c}{DIPVAEI}\\

 Correlation & $s=0.2$ & $s=0.4$  & $s=\infty$  & $s=0.2$ & $s=0.4$  & $s=\infty$ & $s=0.2$ & $s=0.4$  & $s=\infty$& $s=0.2$ & $s=0.4$  & $s=\infty$\\
 \cmidrule{1-1} \cmidrule(lr){2-4} \cmidrule(lr){5-7} \cmidrule(lr){8-10} \cmidrule(lr){11-13}
unit dirs. & \wres{0.666}{0.030} & \wres{0.497}{0.028} & \wres{0.650}{0.049} & \wres{0.441}{0.065} & \wres{0.507}{0.105} & \wres{0.651}{0.087} & \wres{0.580}{0.022} & \wres{0.619}{0.008} & \wres{0.504}{0.056} & \wres{0.686}{0.072} & \wres{0.631}{0.018} & \wres{0.868}{0.052}\\
PCA & \wres{0.287}{0.010} & \wres{0.263}{0.028} & \wres{0.357}{0.024} & \wres{0.312}{0.048} & \wres{0.358}{0.075} & \wres{0.484}{0.064} & \wres{0.341}{0.018} & \wres{0.400}{0.030} & \wres{0.396}{0.061} & \wres{0.266}{0.029} & \wres{0.158}{0.013} & \wres{0.215}{0.037}\\
ICA & \wres{0.394}{0.099} & \wres{0.574}{0.040} & \wres{0.674}{0.012} & \wres{0.193}{0.052} & \wres{0.294}{0.070} & \wres{0.390}{0.109} & \wres{0.516}{0.019} & \wres{0.540}{0.023} & \bwres{0.642}{0.007} & \wres{0.672}{0.073} & \wres{0.630}{0.018} & \bwres{0.870}{0.049}\\
\cmidrule{1-1} \cmidrule(lr){2-4} \cmidrule(lr){5-7} \cmidrule(lr){8-10} \cmidrule(lr){11-13}
Grad (IMA) & \wres{0.638}{0.067} & \wres{0.617}{0.018} & \wres{0.556}{0.109} & \wres{0.478}{0.046} & \wres{0.551}{0.040} & \bwres{0.666}{0.041} & \wres{0.548}{0.035} & \wres{0.623}{0.021} & \wres{0.551}{0.038} & \wres{0.705}{0.062} & \wres{0.644}{0.019} & \wres{0.794}{0.043}\\
IG (IMA) & \wres{0.702}{0.035} & \wres{0.460}{0.128} & \wres{0.578}{0.117} & \wres{0.470}{0.035} & \wres{0.511}{0.042} & \wres{0.581}{0.066} & \wres{0.619}{0.024} & \wres{0.533}{0.006} & \wres{0.612}{0.024} & \wres{0.650}{0.072} & \wres{0.605}{0.006} & \wres{0.701}{0.045}\\
SG (IMA) & \wres{0.677}{0.037} & \wres{0.438}{0.127} & \wres{0.609}{0.131} & \wres{0.475}{0.042} & \wres{0.561}{0.040} & \wres{0.644}{0.055} & \wres{0.533}{0.028} & \wres{0.620}{0.021} & \wres{0.559}{0.040} & \wres{0.698}{0.060} & \wres{0.642}{0.017} & \wres{0.785}{0.046}\\
\cmidrule{1-1} \cmidrule(lr){2-4} \cmidrule(lr){5-7} \cmidrule(lr){8-10} \cmidrule(lr){11-13}
Grad (DMA) & \wres{0.645}{0.067} & \bwres{0.641}{0.031} & \bwres{0.690}{0.062} & \wres{0.547}{0.056} & \wres{0.584}{0.047} & \wres{0.385}{0.169} & \bwres{0.629}{0.033} & \wres{0.666}{0.010} & \wres{0.598}{0.057} & \bwres{0.717}{0.059} & \bwres{0.684}{0.009} & \wres{0.857}{0.037}\\
IG (DMA) & \wres{0.645}{0.076} & \wres{0.530}{0.106} & \wres{0.548}{0.114} & \bwres{0.573}{0.046} & \bwres{0.615}{0.045} & \wres{0.631}{0.128} & \wres{0.607}{0.028} & \wres{0.624}{0.021} & \wres{0.584}{0.039} & \wres{0.703}{0.073} & \wres{0.659}{0.008} & \wres{0.771}{0.029}\\
SG (DMA) & \bwres{0.711}{0.040} & \wres{0.593}{0.094} & \wres{0.633}{0.062} & \wres{0.506}{0.057} & \wres{0.600}{0.027} & \wres{0.644}{0.066} & \wres{0.628}{0.033} & \bwres{0.670}{0.014} & \wres{0.595}{0.059} & \wres{0.716}{0.059} & \wres{0.682}{0.010} & \wres{0.851}{0.036}\\

\bottomrule
\end{tabular}
}
\vspace{0.5em}
\caption{Mean $\pm$ std. err. of the DCI score of the experiments in \cref{sec:exp_comp} for the first correlated component pair (\emph{floor} vs \emph{background} color) in Shapes3D, as an ablation study with further correlations strengths and attribution methods (see \Cref{sec:app_gradtoattrib}). We observe only small differences between attribution methods, with plain Grad and SG performing best in the DMA setting.}
\label{tab:app_further_cor}
\end{table}
\begin{table}[tb]
\adjustbox{width=\textwidth}{
\begin{tabular}{r*{6}{c}}
\toprule
 Dataset & \multicolumn{3}{c}{Shapes3D} & \multicolumn{3}{c}{MPI3D-real}\\
\cmidrule{1-1} \cmidrule(lr){2-4} \cmidrule(lr){5-7}
 \wrapb{Correlated}{factors} & \wrapb{floor vs.}{background} & \wrapb{orientation vs.}{background} &  \wrapb{orientation vs.}{size} &  \wrapb{background vs.}{object color} &  \wrapb{background vs.}{robot arm dof-1} &  \wrapb{robot arm dof-1 vs.}{robot arm dof-2}\\
\cmidrule{1-1} \cmidrule(lr){2-4} \cmidrule(lr){5-7}
\textbf{BetaVAE} &\res{0.309}{0.031}& \res{0.426}{0.043}& \res{0.335}{0.059}&
\res{0.232}{0.022} & \res{0.185}{0.031} & \bres{0.196}{0.034}\\
+PCA&\res{0.111}{0.031}& \res{0.101}{0.009}& \res{0.092}{0.031}& 
\res{0.095}{0.010} & \res{0.105}{0.023} & \res{0.123}{0.033}\\
+ICA&\res{0.360}{0.040}& \res{0.324}{0.054}& \res{0.277}{0.036}& 
\res{0.155}{0.025} & \res{0.163}{0.014} & \res{0.071}{0.014}\\
+Ours (IMA)&\res{0.511}{0.029}& \res{0.437}{0.044}& \res{0.502}{0.030}& 
\bres{0.239}{0.021} & \bres{0.229}{0.022} & \res{0.187}{0.039}\\
+Ours (DMA)&\bres{0.594}{0.023}& \bres{0.485}{0.057}& \bres{0.545}{0.034}& 
\res{0.193}{0.036} & \res{0.092}{0.038} & \res{0.080}{0.015} \\
\cmidrule{1-1} \cmidrule(lr){2-4} \cmidrule(lr){5-7}
\textbf{FactorVAE} &\res{0.297}{0.084}& \res{0.319}{0.076}& \bres{0.423}{0.018}& 
\res{0.079}{0.001} & \res{0.103}{0.020} & \res{0.080}{0.010} \\
+PCA&\res{0.202}{0.057}& \res{0.135}{0.028}& \res{0.235}{0.036}& 
\bres{0.111}{0.006} & \bres{0.122}{0.011} & \bres{0.107}{0.009} \\
+ICA&\res{0.199}{0.061}& \res{0.106}{0.025}& \res{0.078}{0.021}&
\res{0.018}{0.008} & \res{0.061}{0.015} & \res{0.069}{0.015}\\
+Ours (IMA)&\bres{0.337}{0.033}& \bres{0.322}{0.056}& \res{0.288}{0.092}& 
\res{0.070}{0.014} & \res{0.086}{0.018} & \res{0.039}{0.014}\\
+Ours (DMA)&\res{0.276}{0.036}& \res{0.217}{0.064}& \res{0.213}{0.036}& 
\res{0.046}{0.021} & \res{0.045}{0.016} & \res{0.048}{0.015}\\
\cmidrule{1-1} \cmidrule(lr){2-4} \cmidrule(lr){5-7}
\textbf{BetaTCVAE} &\res{0.333}{0.008}& \res{0.400}{0.046}& \res{0.402}{0.017}& 
\bres{0.279}{0.025} & \bres{0.223}{0.030} & \res{0.201}{0.039}\\
+PCA&\res{0.249}{0.033}& \res{0.145}{0.039}& \res{0.184}{0.062}& 
\res{0.265}{0.019} & \res{0.203}{0.028} & \bres{0.213}{0.035}\\
+ICA&\res{0.390}{0.031}& \res{0.276}{0.043}& \res{0.346}{0.072}& 
\res{0.199}{0.040} & \res{0.158}{0.038} & \res{0.170}{0.033}\\
+Ours (IMA)&\res{0.484}{0.025}& \res{0.490}{0.033}& \res{0.526}{0.036}& 
\res{0.092}{0.029} & \res{0.071}{0.029} & \res{0.041}{0.014}\\
+Ours (DMA)&\bres{0.525}{0.014}& \bres{0.540}{0.021}& \bres{0.620}{0.024}& 
\res{0.120}{0.037} & \res{0.122}{0.044} & \res{0.075}{0.028}\\
\cmidrule{1-1} \cmidrule(lr){2-4} \cmidrule(lr){5-7}
\textbf{DipVAE} &\res{0.493}{0.032}& \res{0.481}{0.020}& \res{0.433}{0.044}& 
\res{0.138}{0.020} & \res{0.099}{0.040} & \bres{0.143}{0.045}\\
+PCA&\res{0.063}{0.006}& \res{0.086}{0.027}& \res{0.108}{0.014}& 
\res{0.054}{0.016} & \res{0.042}{0.011} & \res{0.064}{0.010}\\
+ICA&\res{0.495}{0.032}& \res{0.438}{0.053}& \res{0.224}{0.026}& 
\res{0.138}{0.023} & \res{0.096}{0.040} & \res{0.139}{0.047}\\
+Ours (IMA)&\res{0.512}{0.042}& \res{0.425}{0.036}& \res{0.465}{0.049}& 
\bres{0.146}{0.019} & \bres{0.105}{0.033} & \res{0.136}{0.049}\\
+Ours (DMA)&\bres{0.591}{0.028}& \bres{0.546}{0.017}& \bres{0.497}{0.060}& 
\res{0.133}{0.029} & \res{0.094}{0.036} & \res{0.125}{0.045}\\
 \bottomrule
  \end{tabular}
}
\vspace{0.5em}
\caption{Mean $\pm$ std. err. of the Mutual-Information Gap (MIG) scores of four post-hoc methods applied to the embedding spaces of four disentanglement models on two datasets with different pairs of correlated variables. The MIG is computed across all components of the dataset.}
\label{tab:app_posthocdisentangle_mig}
\end{table}

\begin{table}[tb]
\adjustbox{width=\textwidth}{
\begin{tabular}{r*{6}{c}}
\toprule
 Dataset & \multicolumn{3}{c}{Shapes3D} & \multicolumn{3}{c}{MPI3D-real}\\
\cmidrule{1-1} \cmidrule(lr){2-4} \cmidrule(lr){5-7}
 \wrapb{Correlated}{factors} & \wrapb{floor vs.}{background} & \wrapb{orientation vs.}{background} &  \wrapb{orientation vs.}{size} &  \wrapb{background vs.}{object color} &  \wrapb{background vs.}{robot arm dof-1} &  \wrapb{robot arm dof-1 vs.}{robot arm dof-2}\\
\cmidrule{1-1} \cmidrule(lr){2-4} \cmidrule(lr){5-7}
\textbf{BetaVAE} &\bres{0.834}{0.022}& \bres{0.839}{0.053}& \res{0.828}{0.011} & \res{0.557}{0.032} & \res{0.490}{0.044} & \res{0.412}{0.022} \\
+PCA&\res{0.722}{0.060}& \res{0.689}{0.047}& \res{0.716}{0.035}& \res{0.393}{0.037} & \res{0.452}{0.031} & \res{0.398}{0.031}\\
+ICA&\res{0.797}{0.036}& \res{0.775}{0.083}& \res{0.794}{0.022}&
\res{0.385}{0.100} & \res{0.262}{0.061} & \res{0.251}{0.031} \\
+Ours (IMA)&\res{0.767}{0.108}& \res{0.808}{0.060}& \bres{0.832}{0.022}& \res{0.565}{0.022} & \res{0.504}{0.036} & \res{0.443}{0.027} \\
+Ours (DMA)&\res{0.813}{0.087}& \res{0.829}{0.068}& \res{0.826}{0.029}& 
\bres{0.567}{0.024} & \bres{0.525}{0.042} & \bres{0.444}{0.027} \\
\cmidrule{1-1} \cmidrule(lr){2-4} \cmidrule(lr){5-7}

\textbf{FactorVAE} &\res{0.636}{0.045}& \res{0.622}{0.064}& \res{0.595}{0.050}&
\bres{0.354}{0.016} & \bres{0.389}{0.015} & \res{0.342}{0.006}\\
+PCA&\res{0.627}{0.071}& \bres{0.680}{0.027}& \bres{0.652}{0.024}&
\res{0.330}{0.018} & \res{0.388}{0.022} & \bres{0.353}{0.016}\\
+ICA&\res{0.619}{0.059}& \res{0.446}{0.146}& \res{0.200}{0.148}&
\res{0.277}{0.013} & \res{0.242}{0.082} & \res{0.304}{0.017}\\
+Ours (IMA)&\bres{0.663}{0.022}& \res{0.661}{0.028}& \res{0.644}{0.051} &
\res{0.347}{0.007} & \res{0.386}{0.020} & \res{0.337}{0.013}\\
+Ours (DMA)&\res{0.646}{0.026}& \res{0.637}{0.023}& \res{0.619}{0.026}& 
\res{0.330}{0.015} & \res{0.375}{0.016} & \res{0.335}{0.013}\\
\cmidrule{1-1} \cmidrule(lr){2-4} \cmidrule(lr){5-7}

\textbf{BetaTCVAE} &\res{0.676}{0.012}& \res{0.814}{0.052}& \res{0.877}{0.015}& 
\res{0.445}{0.044} & \res{0.379}{0.021} & \res{0.346}{0.020}\\
+PCA&\res{0.761}{0.035}& \res{0.738}{0.063}& \res{0.794}{0.037}&
\bres{0.505}{0.040} & \bres{0.425}{0.012} & \res{0.389}{0.008}\\
+ICA&\res{0.834}{0.004}& \res{0.761}{0.051}& \res{0.806}{0.051}&
\res{0.149}{0.099} & \res{0.168}{0.053} & \res{0.057}{0.035} \\
+Ours (IMA)&\res{0.837}{0.004}& \res{0.849}{0.015} & \bres{0.879}{0.013} & 
\res{0.463}{0.048} & \res{0.401}{0.018} & \res{0.399}{0.019} \\
+Ours (DMA) &\bres{0.842}{0.000} & \bres{0.854}{0.017} & \res{0.878}{0.013} &
\res{0.460}{0.046} & \res{0.399}{0.018} & \bres{0.399}{0.014} \\
\cmidrule{1-1} \cmidrule(lr){2-4} \cmidrule(lr){5-7}

\textbf{DipVAE} &\bres{0.826}{0.006}& \res{0.839}{0.006}& \res{0.785}{0.033}& 
\bres{0.517}{0.046} & \bres{0.473}{0.046} & \res{0.430}{0.013}\\
+PCA&\res{0.671}{0.019}& \res{0.603}{0.064}& \res{0.653}{0.039}& 
\res{0.431}{0.028} & \res{0.373}{0.027} & \res{0.344}{0.021}\\
+ICA&\res{0.826}{0.006}& \res{0.831}{0.007}& \res{0.749}{0.027}&
\res{0.434}{0.042} & \res{0.423}{0.027} & \res{0.424}{0.012}\\
+Ours (IMA)&\res{0.824}{0.007}& \res{0.812}{0.018}& \res{0.785}{0.029}&
\res{0.503}{0.044} & \res{0.471}{0.035} & \res{0.436}{0.021} \\
+Ours (DMA)&\res{0.822}{0.006}& \bres{0.850}{0.012}& \bres{0.809}{0.045}& 
\res{0.505}{0.040} & \res{0.459}{0.040} & \bres{0.448}{0.026}\\
 \bottomrule
  \end{tabular}
}
\vspace{0.5em}
\caption{Mean $\pm$ std. err. of the FactorVAE scores of four post-hoc methods applied to the embedding spaces of four disentanglement models on two datasets with different pairs of correlated variables. The FactorVAE score is computed across all components of the dataset.}
\label{tab:app_posthocdisentangle_factorvae}
\end{table}

\begin{table}[tb]
\adjustbox{width=\textwidth}{
\begin{tabular}{r*{6}{c}}
\toprule
 Dataset & \multicolumn{3}{c}{Shapes3D} & \multicolumn{3}{c}{MPI3D-real}\\
\cmidrule{1-1} \cmidrule(lr){2-4} \cmidrule(lr){5-7}
 \wrapb{Correlated}{factors} & \wrapb{floor vs.}{background} & \wrapb{orientation vs.}{background} &  \wrapb{orientation vs.}{size} &  \wrapb{background vs.}{object color} &  \wrapb{background vs.}{robot arm dof-1} &  \wrapb{robot arm dof-1 vs.}{robot arm dof-2}\\
\cmidrule{1-1} \cmidrule(lr){2-4} \cmidrule(lr){5-7}
\textbf{BetaVAE} &\res{0.086}{0.003}& \res{0.119}{0.004}& \res{0.100}{0.005}&
\res{0.127}{0.014} & \res{0.098}{0.015} & \bres{0.092}{0.025} \\
+PCA&\res{0.047}{0.005}& \res{0.062}{0.006}& \res{0.066}{0.006}& 
\res{0.027}{0.005} & \res{0.055}{0.008} & \res{0.037}{0.006}\\
+ICA&\res{0.007}{0.001}& \res{0.013}{0.001}& \res{0.019}{0.004}&
\res{0.017}{0.006} & \res{0.007}{0.002} & \res{0.004}{0.001} \\
+Ours (IMA)&\bres{0.099}{0.026}& \res{0.114}{0.008}& \res{0.112}{0.007}&
\bres{0.131}{0.011} & \bres{0.113}{0.005} & \res{0.082}{0.024} \\
+Ours (DMA)&\res{0.094}{0.020}& \bres{0.127}{0.012}& \bres{0.114}{0.013}& 
\res{0.107}{0.025} & \res{0.059}{0.024} & \res{0.037}{0.013} \\
\cmidrule{1-1} \cmidrule(lr){2-4} \cmidrule(lr){5-7}

\textbf{FactorVAE} &\res{0.072}{0.006}& \res{0.059}{0.006}& \bres{0.064}{0.001}&
\res{0.059}{0.004} & \res{0.066}{0.008} & \res{0.054}{0.003}\\
+PCA&\res{0.060}{0.006}& \bres{0.066}{0.004}& \res{0.057}{0.004}&
\bres{0.065}{0.008} & \bres{0.076}{0.004} & \bres{0.071}{0.003}\\
+ICA&\res{0.013}{0.002}& \res{0.008}{0.001}& \res{0.006}{0.002}& 
\res{0.002}{0.000} & \res{0.002}{0.001} & \res{0.001}{0.000}\\
+Ours (IMA)&\bres{0.077}{0.012}& \res{0.052}{0.005}& \res{0.054}{0.017}&
\res{0.054}{0.006} & \res{0.059}{0.006} & \bres{0.036}{0.015}\\
+Ours (DMA)&\res{0.071}{0.014}& \res{0.053}{0.012}& \res{0.040}{0.010}& 
\res{0.041}{0.017} & \res{0.043}{0.015} & \res{0.044}{0.013}\\
\cmidrule{1-1} \cmidrule(lr){2-4} \cmidrule(lr){5-7}

\textbf{BetaTCVAE} &\res{0.052}{0.002}& \res{0.107}{0.013}& \res{0.096}{0.016}& 
\bres{0.151}{0.017} & \bres{0.133}{0.007} & \bres{0.117}{0.011}\\
+PCA&\res{0.073}{0.004}& \res{0.075}{0.011}& \res{0.107}{0.015}&
\res{0.148}{0.018} & \res{0.125}{0.009} & \res{0.109}{0.007}\\
+ICA&\res{0.015}{0.000}& \res{0.010}{0.001}& \res{0.011}{0.002}& 
\res{0.011}{0.004} & \res{0.005}{0.002} & \res{0.004}{0.002}\\
+Ours (IMA)&\res{0.105}{0.003}& \res{0.119}{0.012}& \bres{0.130}{0.023}&
\res{0.055}{0.017} & \res{0.059}{0.016} & \res{0.056}{0.003} \\
+Ours (DMA)&\bres{0.108}{0.005}& \bres{0.127}{0.013}& \res{0.109}{0.017}& 
\res{0.071}{0.020} & \res{0.072}{0.010} & \res{0.051}{0.015} \\
\cmidrule{1-1} \cmidrule(lr){2-4} \cmidrule(lr){5-7}

\textbf{DipVAE} &\res{0.083}{0.004}& \res{0.084}{0.003}& \res{0.070}{0.002}&
\res{0.056}{0.011} & \res{0.039}{0.013} & \res{0.057}{0.016}\\
+PCA&\res{0.027}{0.003}& \res{0.034}{0.006}& \res{0.043}{0.004}& 
\res{0.023}{0.004} & \res{0.030}{0.008} & \res{0.022}{0.005}\\
+ICA&\res{0.006}{0.001}& \res{0.003}{0.002}& \res{0.030}{0.002}&
\res{0.011}{0.005} & \res{0.005}{0.003} & \res{0.005}{0.002}\\
+Ours (IMA)&\res{0.089}{0.012}& \res{0.082}{0.005}& \res{0.077}{0.002}&
\bres{0.060}{0.008} & \bres{0.047}{0.010} & \bres{0.061}{0.016} \\
+Ours (DMA)&\bres{0.114}{0.003}& \bres{0.105}{0.008}& \bres{0.084}{0.007}&
\res{0.051}{0.008} & \res{0.043}{0.012} & \res{0.054}{0.016}\\
 \bottomrule
  \end{tabular}
}
\vspace{0.5em}
\caption{Mean $\pm$ std. err. of the SAP scores of four post-hoc methods applied to the embedding spaces of four disentanglement models on two datasets with different pairs of correlated variables. The SAP score is computed across all components of the dataset.} 
\label{tab:app_posthocdisentangle_sap}
\end{table}

\subsection{Qualitative results on Shapes3D}
In this section, we want to show another traversal plot like the one in \cref{fig:travsersal} and more thoroughly analyze its latent space. We chose another architecture (BetaTCVAE) and $s=0.2$ with the usual correlated factors \emph{floor color} and \emph{background color}. Out of the 5 independent runs, we selected the one with the highest DCI score (of the base model) for the analysis.
\newcommand{\includetraversalsupp}[1]{\includegraphics[clip, trim=0.37cm 2.24cm 0.3cm 1.7cm, width=0.75\textwidth]{#1}}
\newcommand{\includetraversalsuppB}[1]{\includegraphics[clip, trim=0.37cm 2.24cm 0.3cm 1.7cm, width=0.75\textwidth]{#1}}
\newcommand{\includematrixplot}[1]{\includegraphics[clip, trim=0.1cm 1.6cm 0.3cm 0.4cm, width=0.65\textwidth]{#1}}
\newcommand{\includematrixplotB}[1]{\includegraphics[clip, trim=0.1cm 0.2cm 0.3cm 0.4cm, width=0.75\textwidth]{#1}}
\ifdefined\arxiv
\newcommand{\sfigw}[0]{0.4\textwidth}
\else
\newcommand{\sfigw}[0]{0.48\textwidth}
\fi
\begin{figure}[tb]
   \begin{subfigure}[b]{\sfigw}
    \centering
    \includetraversalsuppB{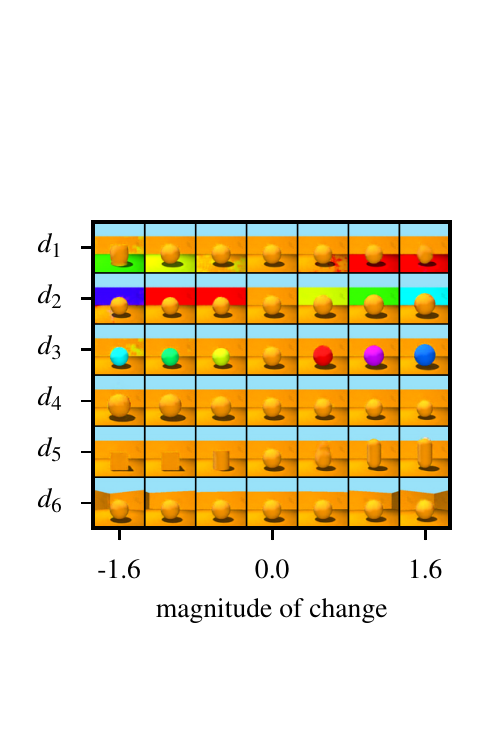}
    \caption{optimal linear directions (traversal plot)\label{fig:app_travlin}}
    \end{subfigure}
    \begin{subfigure}[b]{\sfigw}
    \centering
    \includematrixplotB{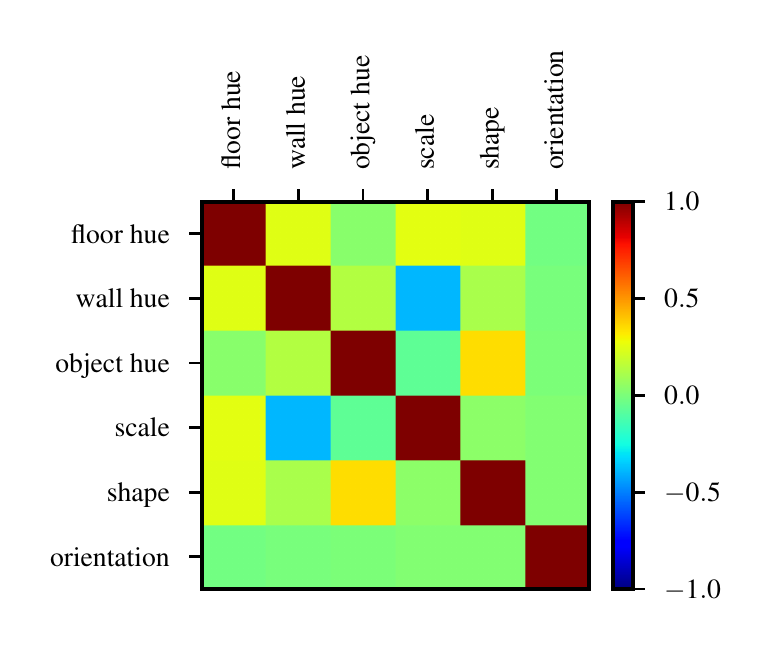}
    \caption{Cosines of the linear directions\label{fig:app_orthmatrix}}
    \end{subfigure}
    \caption{Empirical results for linear entanglement. For the model shown in \cref{fig:travsersal} (trained on correlated data), we observe almost perfect linear entanglement, i.e., that $f \circ g = D$: (a) There exist linear directions $d_1$ to $d_6$ in $f$'s embedding space that encode the individual components. (b) However, these directions are not necessarily orthogonal; they can be entangled as testified by non-zero cosine distances between them. See \cref{fig:app_traversal} for additional results.}
\end{figure}
\paragraph{Linear entanglement matrix.} To study which factors are encoded in which latent dimension, we compute a matrix of linear entanglement. By our linear entanglement hypothesis, $\vz' = \mD\vz$, where the matrix $\mD=\lbrack \vd_1, \cdots, \vd_K\rbrack \in \mathbb{R}^{K\times K}$ contains the directions $\vd_i \in \mathbb{R}^K$, in which the ground truth concepts are encoded. Changing the component $i$ (entry $\vz_i$) by one unit will change the resulting embedding by $\vd_i$. To find these $\vd_i$, we take the factors at the origin of the traversal plot and alter only a single component $i$. We then encode the image corresponding to that change, and measure the change in embeddings to find the linear direction $\vd_i$ that the corresponding component is encoded in (to be precise, we sample several changes and take the largest eigenvector of the embedding changes covariance). Thus, we can estimate the matrix $\mD$. An example is shown in \cref{fig:app_travlin} and provides evidence that linear entanglement is possible when training autoencoder models from correlated data.

To estimate which factors are changing when a unit direction of the (plain or post-processed) embedding space is followed (a change in $\vz_i^{'}$), we can invert the equation to $\vz=\mD^{-1}\vz'$. The columns in $\mD^{-1}$ correspond to the change in ground truth components that going one unit in the latent space coordinate $i$ will entail. We refer to this matrix $\mD^{-1}$, that shows which ground truth components will be altered by moving along one latent dimension as \emph{linear entanglement matrix}.

\begin{figure}[tb]
   \begin{subfigure}[b]{\sfigw}
    \centering
    \includetraversalsupp{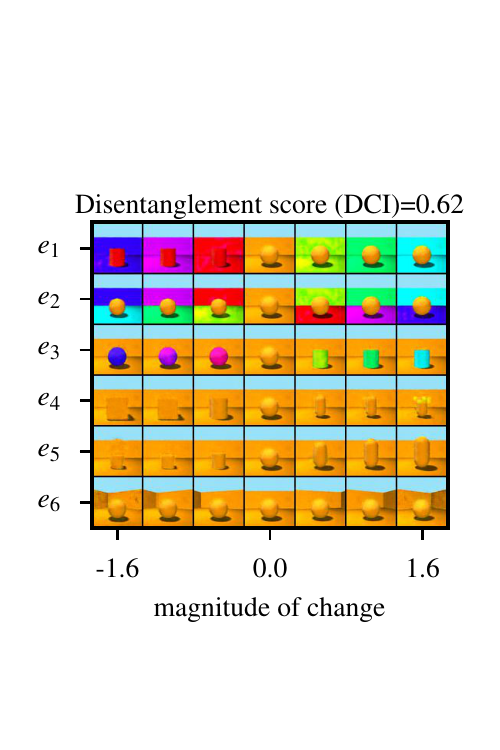}
    \caption{BetaTCVAE \label{fig:travunittc}}
    \end{subfigure}
    \begin{subfigure}[b]{\sfigw}
    \centering
    \includematrixplot{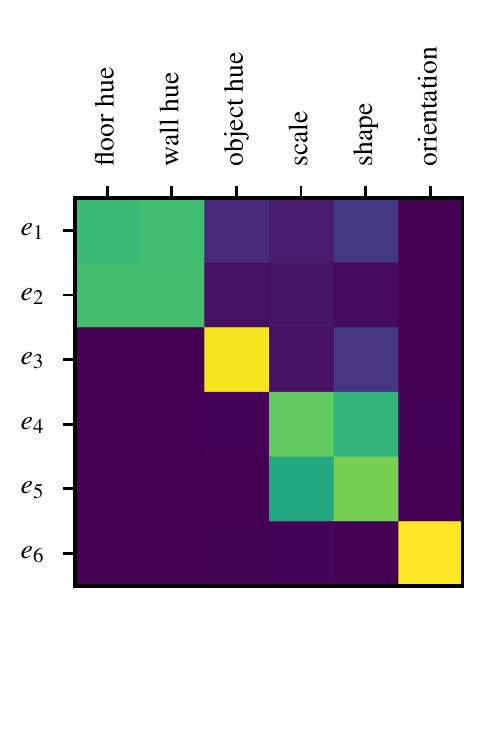}
    \caption{Corresponding linear entanglement matrix\label{fig:matrixunittc}}
    \end{subfigure}
   \begin{subfigure}[b]{\sfigw}
    \centering
    \includetraversalsupp{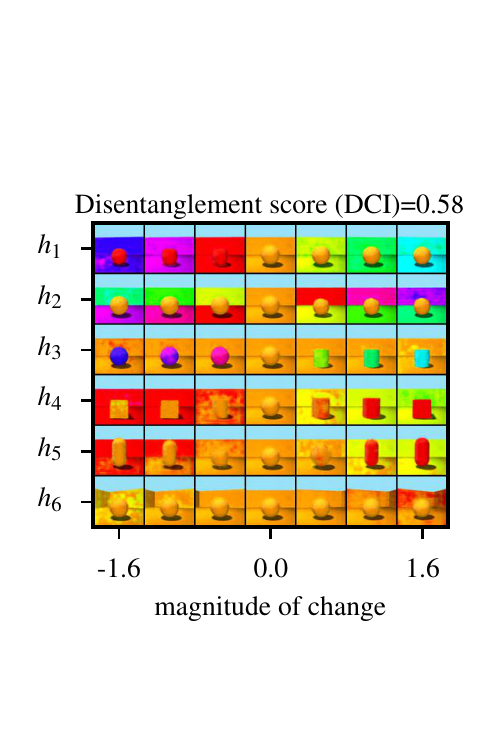}
    \caption{BetaTCVAE + ICA \label{fig:travunittc2}}
    \end{subfigure}
    \begin{subfigure}[b]{\sfigw}
    \centering
    \includematrixplot{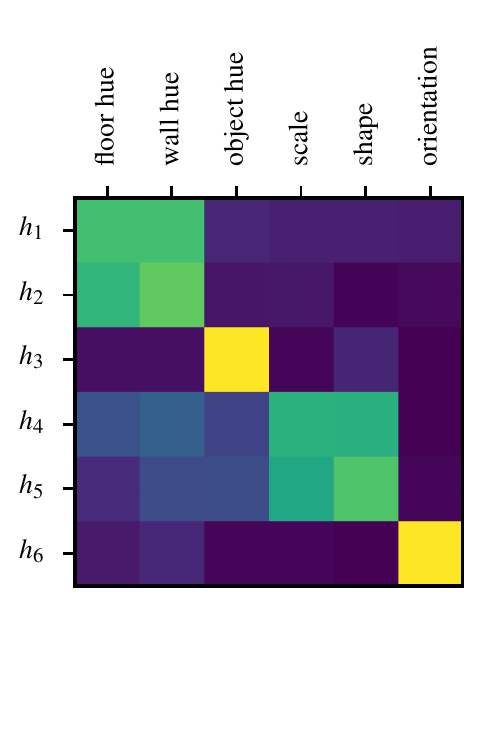}
    \caption{Corresponding linear entanglement matrix\label{fig:matrixunittc2}}
    \end{subfigure}
    \begin{subfigure}[b]{\sfigw}
    \centering
    \includetraversalsupp{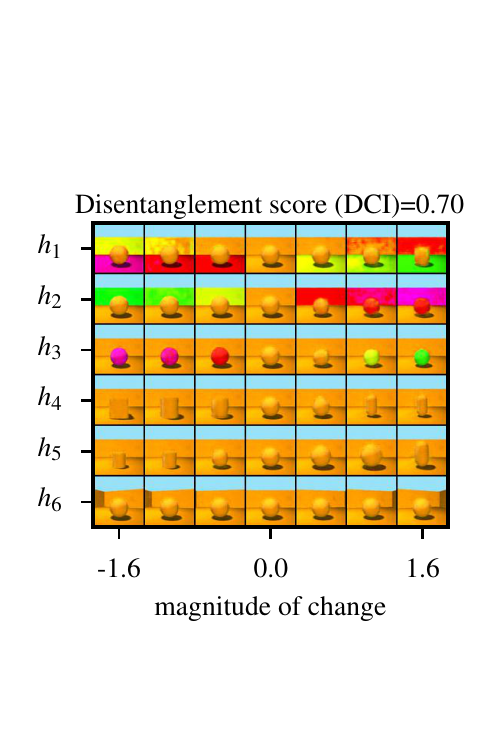}
    \caption{BetaTCVAE + Ours (DMA, SG) \label{fig:travunittc3}}
    \end{subfigure}
    \hfill
    \begin{subfigure}[b]{\sfigw}
    \centering
    \includematrixplot{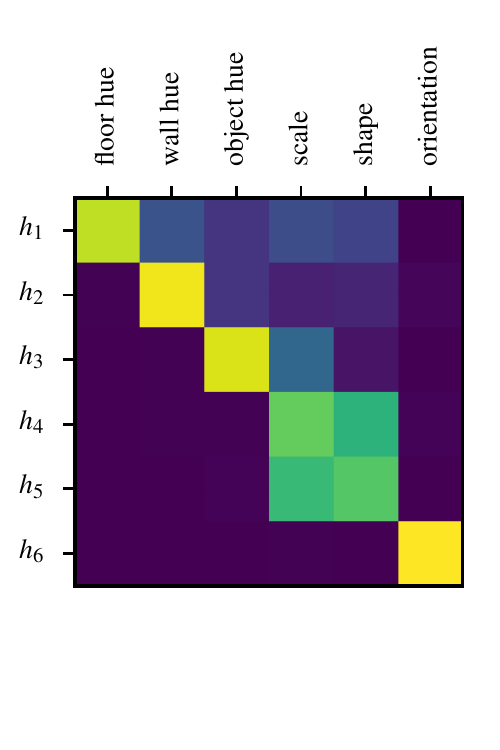}
    \caption{Corresponding linear entanglement matrix\label{fig:matrixunittc3}}
    \end{subfigure}
    \caption{Traversal plots from another model (BetaTCVAE) trained on the correlated dataset. As for all traversal plots in this paper, we manually permuted the dimensions to match across plots. In addition, we compute a matrix of linear entanglement that shows which ground truth factors is changed when moving into a certain direction (brightness corresponds to maginitude of change). While none of the post-hoc methods manages to disentangle shape and size (most likely due to their non-linear encoding), our model resolves the linearly entangled factors \emph{floor hue} and \emph{wall hue} fairly well, which can also be seen from the entanglement matrix. \label{fig:app_traversal}}
\end{figure}

\Cref{fig:app_traversal} shows the traversals along with the corresponding linear entanglement matrices that correspond well to the changes observed. For the plain method, the components that were correlated are deeply entangled (upper line). However, our method (DMA, SG, lower line) is able to separate them well, which is testified both by the traversal and the linear disentanglement matrix. 

\subsection{Further results on CUB} 
\label{sec:app_cub_eval}
For a quantitative evaluation, we match the discovered concepts on CUB with the annotated ground truth attributes. we report results for the quantitative comparison on CUB introduced in \cref{sec:app_cubdetails} of our methods with PCA, ICA, and a baseline of randomly sampled directions.  Furthermore, we implement ConceptSHAP \citep{yeh2019completeness} and ACE \citep{ghorbani2019towards} and use them to discover concepts on CUB (using their default settings otherwise).
The results of this metric are presented in \Cref{tab:cub_quantitative}.

\begin{figure}[tb]
    \centering
    \includegraphics[width=.5\linewidth]{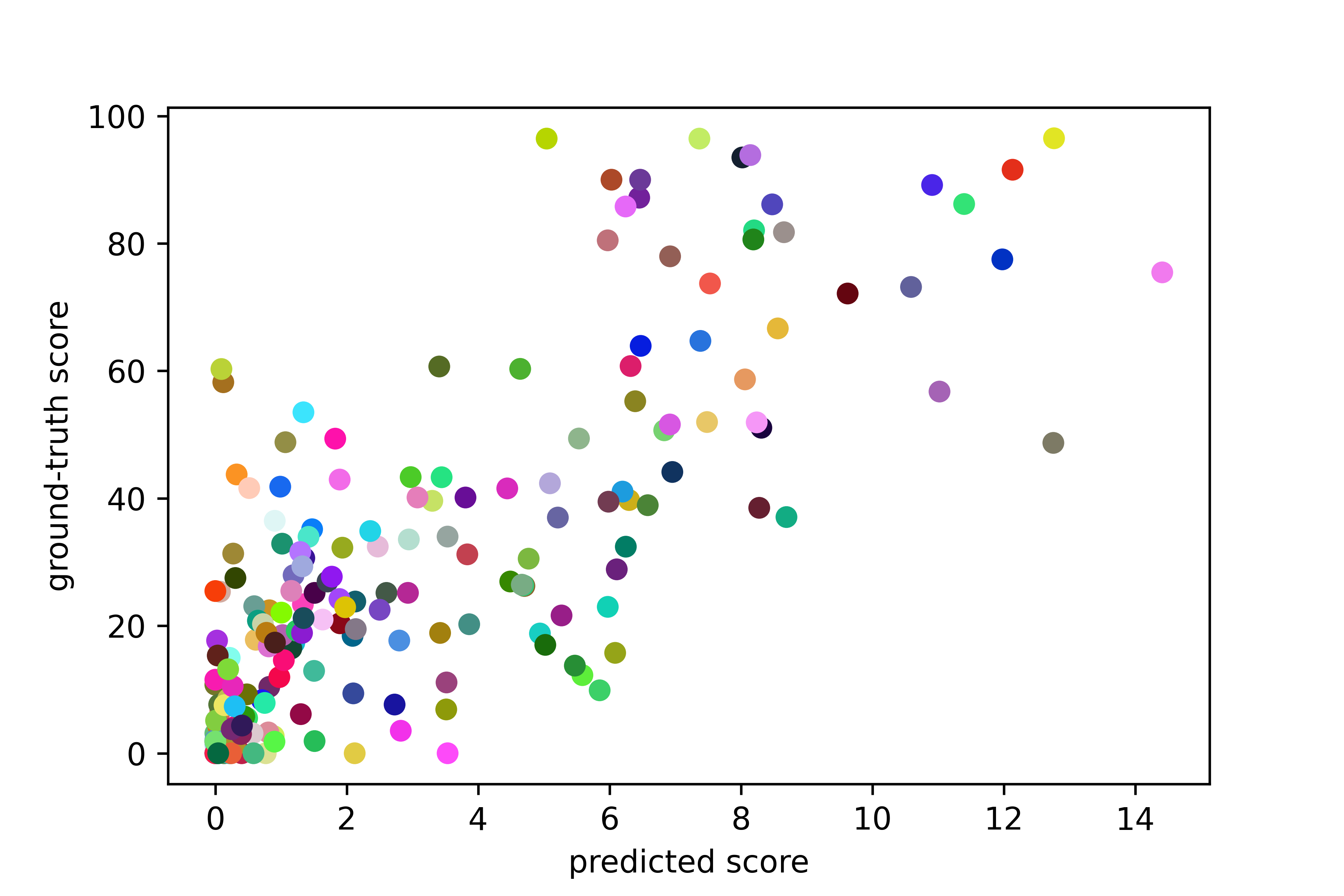}
    \caption{Correlation between ground-truth attribute scores and our predicted scores for the best matched component. Each dot represents a class.}
    \label{fig:cub_correlation}
\end{figure}

\begin{figure}[tb]
    \centering
    \input{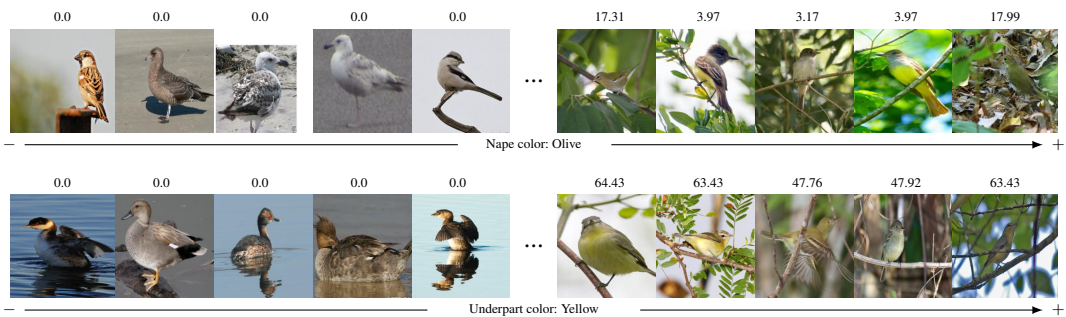}
\caption{Examples of discovered components on CUB. The corresponding ground-truth attribute is shown under images and the ground-truth value of each image is depicted above the image. ``$+$/$-$'' indicate the positive/negative direction along the discovered concept.}
    \label{fig:cub_attr2}
\end{figure}

ICA failed to discover meaningful components, while PCA was only capable of discovering very few high-variance ones in the beginning, but begins to fail for $K>10$. This is possibly because in PCA, the directions are required to be orthogonal. Surprisingly, both PCA and ICA were not much better than the random baseline.  Regarding ConceptSHAP and ACE, we find that ACE often focused on the background concepts and ConceptSHAP discovered concepts that are usually more focused on the birds but hard to localize in a fine-grained manner. Our method constantly discovered components and surpassed all three baselines. 
In particular, our method (DMA) lead to good performance. 
This leads us to the hypotheses that for high-dimensional data, the disjointness principle is required to identify solutions. 
\Cref{fig:cub_correlation} illustrates the correlation between the ground-truth attribute representation (scores) and predicted representation by using our model (using plain gradients) for the top discovered component. The two components are clearly correlated, but more in a block-sense: Classes with low scores on the attribute received low scores on the discovered component. The same holds for high scores, but within these, we observe stronger noise, which explains why the Spearman's correlation values were imperfect. This can be due to a certain degree of arbitrage in the ground-truth attribute values of each class. Here, \cref{fig:cub_attr2}, just like \cref{fig:cub_attr} in the main paper, shows qualitative examples, including the ground-truth values which appear to fluctuate. We emphasize that this analysis should be viewed as an initial take on quantifying the quality of interpretable components, but that a refined benchmark is material for future work.

\begin{table}[tb]
\centering
\resizebox{0.8\linewidth}{!}{
\begin{tabular}{rccccccc}
\toprule
Num. components & K=1 & K=10 & K=20 & K=30 \\ 
\midrule
PCA & \textbf{0.789}  $\pm$ 0.024  &  0.602 $\pm$ 0.007 & 0.497 $\pm$  0.005 &  0.440 $\pm$ 0.006 \\ 
ICA &  0.515 $\pm$ 0.028  & 0.442 $\pm$ 0.005 & 0.412 $\pm$ 0.006  &  0.390 $\pm$ 0.007  \\ 
\midrule
ACE \citep{ghorbani2019towards} & 0.623 $\pm$ 0.012 & 0.579 $\pm$ 0.010 & 0.550 $\pm$ 0.008 & 0.527 $\pm$ 0.007 \\
ConceptSHAP \citep{yeh2019completeness} & 0.655 $\pm$ 0.014 & 0.596 $\pm$ 0.006 & 0.568 $\pm$ 0.008 & 0.545 $\pm$ 0.006 \\
\midrule
Ours-IMA,Grad  & 0.657 $\pm$  0.025 & 0.601 $\pm$ 0.009  & 0.564 $\pm$ 0.009   & 0.535 $\pm$ 0.008 \\
Ours-DMA,Grad      & 0.701  $\pm$  0.045 & \textbf{0.626} $\pm$ 0.029  & \textbf{0.585} $\pm$ 0.028   & \textbf{0.559} $\pm$ 0.011 \\
\bottomrule
\end{tabular}
}
\vspace{0.5em}
\caption{Quantitative comparison of discovered components using our methods, PCA, ICA and a random baseline. Mean correlation score of top-K (K in column) discovered components are shown in (mean ± std.) for five runs.}
\label{tab:cub_quantitative}
\end{table}

\clearpage
\newpage
\begin{@fileswfalse}
\bibliography{references}
\end{@fileswfalse}

\end{document}